\documentclass[10pt]{article} 
\usepackage[preprint]{rlc}

\usepackage{amssymb}      
\usepackage{mathtools}          
\usepackage{mathrsfs}           
\mathtoolsset{showonlyrefs}     
\usepackage{graphicx}           
\usepackage{subcaption}         
\usepackage[space]{grffile}     
\usepackage{url}       
\usepackage{lipsum}
\usepackage{xcolor}
\usepackage{amsthm}
\usepackage{wrapfig}
\usepackage{float}
\usepackage{algorithm}
\usepackage{algpseudocode}
\usepackage{multirow}
\usepackage{xspace}
\usepackage{tikz,ifthen}

\definecolor{lightgrey}{rgb}{.99,.99,.99}
\definecolor{myred}{rgb}{.76,.07,.12}
\definecolor{lapisblue}{rgb}{.15,.38,.61}

\usetikzlibrary{arrows,trees,backgrounds,automata,shapes,decorations,plotmarks,fit,calc,positioning,shadows,chains}
\definecolor{color0}{RGB}{239, 48, 84} 
\definecolor{color4}{RGB}{62, 180, 139} 
\definecolor{color6}{RGB}{147, 129, 255} 
\definecolor{color7}{RGB}{242, 193, 78} 

\definecolor{nnedgecolor}{RGB}{90,90,90}
\tikzstyle{every pin edge}=[<-,shorten <=1pt]
\tikzstyle{every path}=[draw=color7!50]
\tikzstyle{neuron}=[circle,fill=black!25,minimum size=17pt,inner sep=0pt]
\tikzstyle{input neuron}=[neuron, fill=color4]
\tikzstyle{output neuron}=[neuron, fill=red!50]
\tikzstyle{hidden neuron}=[neuron, fill=blue!50]
\tikzstyle{annot} = [text width=4em, text centered]
\tikzstyle{nnedge} = [-{stealth},shorten >=0.1cm, shorten <=0.05cm,line 
width=0.8pt,nnedgecolor]
\usetikzlibrary{calc}

\newcommand{\sat}{\texttt{SAT}}
\newcommand{\turnLeft}{\texttt{TURN LEFT}}
\newcommand{\turnRight}{\texttt{TURN RIGHT}}
\newcommand{\collide}{\texttt{COLLIDE}}
\newcommand{\unsat}{\texttt{UNSAT}}
\newtheorem{definition}{Definition}
\newcommand{\mapless}{\texttt{Mapless Navigation}}
\newcommand{\gridenv}{\texttt{Particle World}}

\newcommand\blfootnote[1]{%
  \begingroup
  \renewcommand\thefootnote{}\footnote{#1}%
  \addtocounter{footnote}{-1}%
  \endgroup
}

\title{Verification-Guided Shielding for Deep \\ Reinforcement Learning}

\author{%
Davide Corsi$^{1, *}$, Guy Amir$^{2, *}$, Andoni Rodríguez$^{3,4}$, \\
\textbf{César Sánchez}$^{3}$, \textbf{Guy Katz}$^2$ \textbf{and Roy Fox}$^{1}$ \\ \\
$^1$University of California, Irvine, USA \\
$^2$The Hebrew University of Jerusalem, Israel\\
$^3$IMDEA Software Institute, Spain\\
$^4$Universidad Politécnica de Madrid, Spain\\
}

\usepackage{ltlfonts}

\newcommand{\AP}{\ensuremath{\mathsf{AP}}}

\newcommand{\DefOR}{\ensuremath{\hspace{0.2em}\big|\hspace{0.2em}}}

\newcommand{\Always}{\LTLsquare}
\newcommand{\Event}{\LTLdiamond} 
\newcommand{\Next}{\LTLcircle}

\newcommand{\U}{\mathbin{\mathcal{U}}}

\newcommand{\Or}{\mathrel{\vee}}

\newcommand{\mycal}[1]{\ensuremath{\mathcal{#1}}}

\newcommand{\calR}{\mycal{R}}

\definecolor{darkGray}{gray}{0.55} 
\definecolor{lightGray}{gray}{0.85} 

\newcommand{\true}{\top}

\begin{document}

\maketitle

\begin{abstract}
In recent years, Deep Reinforcement Learning (DRL) has emerged as an effective approach to solving real-world tasks. However, despite their successes, DRL-based policies suffer from poor reliability, which limits their deployment in safety-critical domains. Various methods have been put forth to address this issue by providing formal safety guarantees. Two main approaches include \emph{shielding} and \emph{verification}. While shielding ensures the safe behavior of the policy by employing an external online component (i.e., a ``shield'') that overrides potentially dangerous actions, this approach has a significant computational cost as the shield must be invoked at runtime to validate \emph{every} decision. On the other hand, verification is an offline process that can identify policies that are unsafe, prior to their deployment, yet, without providing alternative actions when such a policy is deemed unsafe. In this work, we present \emph{verification-guided shielding} --- a novel approach that bridges the DRL reliability gap by integrating these two methods. Our approach combines both formal and probabilistic verification tools to partition the input domain into safe and unsafe regions. In addition, we employ clustering and symbolic representation procedures that compress the unsafe regions into a compact representation. This, in turn, allows to \emph{temporarily} activate the shield solely in (potentially) unsafe regions, in an efficient manner. Our novel approach allows to \emph{significantly} reduce runtime overhead while still preserving formal safety guarantees. We extensively evaluate our approach on two benchmarks from the robotic navigation domain, as well as provide an in-depth analysis of its scalability and completeness. 
\blfootnote{$^*$\textit{Both authors contributed equally.}}
\end{abstract}


\section{Introduction}
\label{sec:introduction}

Deep reinforcement learning (DRL) is gaining popularity due to its recent success in solving complex decision-making problems across various domains and settings \citep{MnKaSi13, kober2013reinforcement, rolf2023review, karamzade2024reinforcement}. However, upon formal and rigorous analysis, even policies generated by state-of-the-art algorithms exhibit a significant drawback: they can not ensure the correctness of the DRL policy for every possible input~\citep{KaHuIbJuLaLiShThWuZeDiKoBa19, CoMaFa21}. This limitation hinders the full integration of DRL agents in safety-critical scenarios, such as autonomous navigation systems \citep{TaPaLi17}, robotic controllers \citep{aractingi2023controlling}, healthcare \citep{pore2021safe}, and decision support in regulated industries \citep{singh2022reinforcement} in which even a single mistake can have dire consequences. This setback emphasizes the need for ensuring the absolute compliance of DRL policies with user-specified safety and behavioral requirements \citep{RaAcAm19, ma2024learn}. 
To this end, the DRL community has recently made significant efforts to generate more reliable agents. These attempts include online training methods such as constrained optimization~\citep{ yang2022constrained, zhang2020first} and safe exploration~\citep{simao2021alwayssafe, kamran2022modern}. However, despite promising results, these approaches are heuristic in nature and are unable to guarantee the absolute correctness of the DRL policy in question. This limitation is observed even when the policy is generated with state-of-the-art algorithms, and for performing relatively simple tasks \citep{corsi2024analyzing}.


An alternative family of approaches tackles the DRL safety problem from a different perspective, by decoupling the safety requirements from the training procedure. These techniques typically involve a formal analysis of the neural network function or the integration of various types of domain expert knowledge into the policy. 
Two of the most promising approaches in this context are formal verification~\citep{LiArLaBaKo19} and shielding~\citep{BlKoKoWa15}. Unlike training-based methods, these techniques are indeed able to provide rigorous guarantees, but they suffer from various limitations. Formal verification, for example, is computationally hard~\citep{KaBaDiJuKo17}, and its applicability to various use cases is thus limited (e.g., it is not clear how to verify large language models). In addition, formal verification tools typically return a binary answer, indicating whether the safety requirement holds or not, without providing any alternative solution when the latter occurs, and the policy is deemed unsafe.
On the other hand, shielding techniques introduce an external component (i.e., a ``shield''), that can override the original unsafe decisions, hence providing a safe action when encountering an unsafe input.
However, although shielding affords safety certifications, there is still no guarantee that the proposed action is optimal~\citep{AlBloEh18} and, crucially, the external shield must be invoked in every time step, resulting in significant overhead. This issue is critical, because in many real-time applications such an overhead may be infeasible in practice.

In this work, we begin bridging this gap, and present \emph{verification-guided shielding}, a novel method that combines both these aforementioned techniques. 
Our approach consists of two main stages. First, we employ a combination of different formal methods to identify all the regions in the input space where the agent is guaranteed to behave correctly. In these regions, we can rely on the original policy, without invoking the external shield to validate the agent's decisions (we note that this is possible only due to the rigorous guarantee provided by the formal verification process). Then, in the remaining (unsafe) input region, we activate the shield, which can potentially override the unsafe decision, when encountered. 
Our approach significantly reduces the overall overhead of ``traditional shielding'', while still preserving the formal guarantees regarding the policy's safety.
Implementing our approach poses several challenges, ranging from scalability to the soundness of the algorithm, which we thoroughly analyze in the following sections. Finally, to demonstrate the effectiveness of our approach, we extensively evaluate it on two popular DRL benchmarks: (i) \gridenv{}, where an agent is trained to navigate in a two-dimensional grid, and (ii) \mapless{}, a real-world task in robotics, where a robot learns to navigate in an unknown arena and reach a given target~\citep{pore2021safe, CoMaFa21}. 
%
%
We use expressive shields for such tasks~\citep{RoAmCoSaKa24}.

The rest of the paper is organized as follows. Sec.~\ref{sec:preliminaries} contains background on safe DRL, formal verification, and shielding. We formalize our problem in Sec.~\ref{sec:motivation}. In Sec.~\ref{sec:method}, we present our novel method for verification-guided shielding, and empirically evaluate it in Sec.~\ref{sec:evaluation}. Related work is covered in Sec.~\ref{sec:related-work}, and we conclude in Sec.~\ref{sec:conclusion}. 
\section{Preliminaries}
\label{sec:preliminaries}

Deep reinforcement learning algorithms typically aim to optimize the \textit{expected cumulative reward}, which represents the main objective of the agent~\citep{sutton2018reinforcement}. However, in safety-critical tasks, it is common to introduce an additional function that represents the safety constraints that should be met as part of the optimization process. Finding a successful policy under these multiple objectives has emerged as a challenging problem~\citep{ma2024learn}. Moreover, it is important to note that DRL training algorithms are designed to fulfill requirements only in expectation, without any formal guarantee on the behavior of the policy during deployment.

\subsection{Formal Verification}
In recent years, various methods have been put forth to formally verify the correctness of deep neural networks (DNNs). These approaches \emph{rigorously} verify whether a given DNN adheres to a safety specification, for \emph{every possible input}. More formally, the DNN verification problem~\citep{KaBaDiJuKo17} is defined as follows: 

\begin{definition}[\textit{The DNN-Verification Problem}]
\label{def:decision_problem}
\phantom{a}

    \fbox{\parbox{\dimexpr\linewidth-2\fboxsep-2\fboxrule}{
    {\bf Input}: $\mathcal{R}=\langle\mathcal{N}, \mathcal{P}, \mathcal{Q}\rangle$, where $\mathcal{N}$ is a DNN, $\mathcal{P}$ is a precondition on the DNN's inputs, and $\mathcal{Q}$ is a postcondition on the DNN's outputs.
    
    {\bf Output}: \sat{} if $\exists\;x\;|\;\mathcal{P}(x) \wedge \mathcal{Q}(\mathcal{N}(x))$, and \unsat{} otherwise.
    }}
\end{definition}

The precondition $\mathcal{P}$ usually encodes domain-specific knowledge on the input space, e.g., it can limit the inputs to represent a specific dangerous situation. The postcondition $\mathcal{Q}$ encodes the \emph{negation} of the desired behavior when the agent's current state belongs to $\mathcal{P}$. Hence, when a verification algorithm (the ``verifier'') answers \unsat, i.e., that there does not exist a satisfying assignment, this indicates that the DNN behaves correctly on all inputs in our domain of interest. On the other hand, when the verification algorithm returns \sat{}, this indicates that a satisfying assignment is found, and at least a single input $x$ adheres to $\mathcal{P}(x) \wedge \mathcal{Q}(\mathcal{N}(x))$, and triggers the unwanted behavior. 
The DNN verification problem is computationally hard and has been proven to be NP-complete~\citep{KaBaDiJuKo17}, hence, such techniques are usually applied only in safety-critical tasks, in which the safety of the DRL in question must be rigorously guaranteed, and classic testing techniques are inadequate.


\textbf{DNN Verification Example}. 
DNN verification can be employed in many real-world problems. For instance, it has been shown that DNNs are susceptible to \emph{adversarial inputs}, i.e., small input perturbations that can cause even the best DNNs to fail miserably~\citep{SzZaSuBrErGoFe13, HuPaGoDuAb17, MaDiMe20, FeYi20, GoLiZhSaYuLiWaFe20}. The resilience, or \emph{robustness}, of a DNN to such perturbations can directly be assessed using off-the-shelf verifiers~\citep{TjXiTe17, 
ZhWeChHsDa18, GoKaPaBa18, CaKoDaKoKaAmRe22} by encoding as a precondition an $\epsilon$-ball around a given input $x$ ($\mathcal{P}(x)\coloneqq x\in B_{\epsilon}(x)$), and as a postcondition a case in which the DNN \emph{misclassifies} a given input $x'\in B_{\epsilon}(x)$. 
In the context of deep reinforcement learning, the verified properties are typically safety constraints, that are also encoded (by domain experts) as input-output relations. For additional details, see Appendix~\ref{sec:app:verification-properties}.

\subsection{LTL Synthesis and Shielding}

Linear temporal logic (LTL) is a type of logic pertaining to modalities referring to linear time~\citep{Pn77,MaPn95}. %
In LTL, it is possible to encode formulae regarding the various states and actions throughout multiple time-steps, e.g., \emph{there are no three consecutive states in which a given action is chosen}.
%
More formally, the LTL syntax is recursively defined as follows:
\[
  \varphi  ::= \true \DefOR a \DefOR \varphi \lor \varphi \DefOR \neg \varphi
  \DefOR \Next \varphi \DefOR \Always \varphi \DefOR \varphi \U\varphi, 
\]
where  $a\in\AP$ is an \emph{atomic proposition}, $\{\land,\neg\}$ are the common Boolean operators of \emph{conjunction} and \emph{negation}, respectively, and $\{\Next,\U,\Always\}$ are the \emph{next}, \emph{until} and \emph{always} temporal operators, respectively. 
Reactive LTL synthesis~\citep{PiPnSa06,Th08} is the task of automatically producing a system that satisfies a given LTL specification $\varphi$, where atomic propositions in $\varphi$ are split into variables assigned by an uncontrollable environment (input variables $I$) and variables assigned by a controllable system (output variables $O$). We refer the reader to Appendix~\ref{sec:app:synthesis} for an in-depth description of LTL semantics and synthesis.

\textbf{DRL Shielding.} Recently, it has been shown that a given LTL formula $\varphi$ represents a desired specification that can be used to automatically synthesize shields~\citep{BlKoKoWa15,AlBloEh18}, i.e., generate external components that are coupled with the agent, and \emph{force} it to behave safely according to the specification $\varphi$. More formally, given an LTL specification $\varphi$, 
\begin{wrapfigure}{r}{0.5\textwidth}
    \vspace{-10pt}
    \begin{center}
        \includegraphics[width=0.4\textwidth]{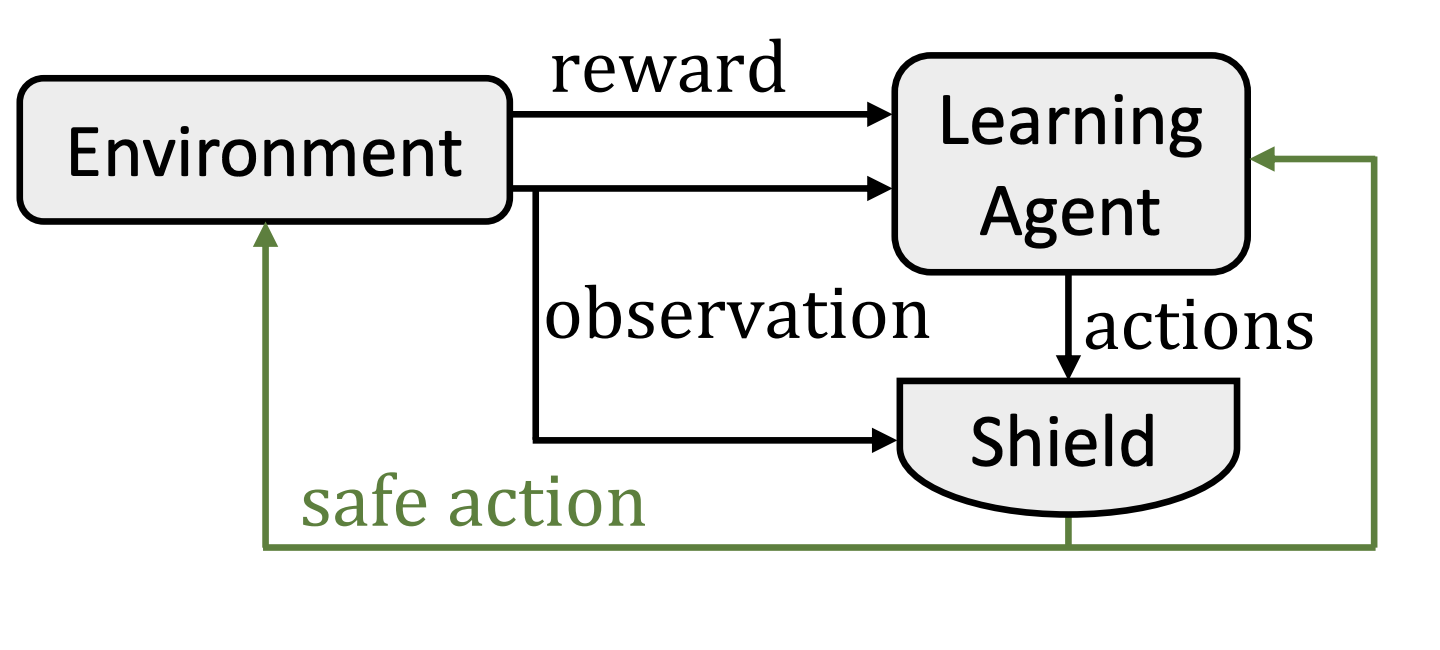}
    \end{center}
    \vspace{-10pt}
    \caption{A shielding architecture scheme for a DRL agent~\citep{AlBloEh18}.
    }
    \label{fig:ShieldBasic}
    \vspace{-10pt}
\end{wrapfigure}it is possible to generate a shield $S$ with respect to a given system $D$ (controlled by a DRL agent, in our case). $S$ guarantees that \emph{all} behaviors of the DRL-controlled system $D$ satisfy $\varphi$ as follows: when $S$ encounters an input $I$ that triggers an erroneous output (i.e., $D(I)\coloneqq O$ for which $\varphi(I,O)$ does not hold), the original action $O$ is corrected, and replaced with another action $O'$, ensuring that $\varphi$(I, O') does hold. This scheme, depicted in Fig.~\ref{fig:ShieldBasic}, guarantees that the combined system  $D \circ S$ never violates $\varphi$. 

\textbf{Example.} Given the atomic proposition $\collide$ (indicating that the agent collided), the LTL formula $\phi \coloneqq \Always (\neg \collide)$ encodes the safety property in which for all steps (``always''), no collision occurs. Given this requirement, and given that a DRL-controlled agent $D$ observes an input $I$ representing an obstacle to the left --- an action $O \coloneqq \turnLeft$ will cause a collision in the next step (hence violating $\phi$). Thus, a shield $S$ may override $O$ with an alternative action (e.g., $O'\coloneqq \turnRight$), satisfying $\phi$ and hence maintaining safety. 

\begin{figure}[b]
\centering
\minipage{0.22\linewidth}
    \includegraphics[width=\linewidth]{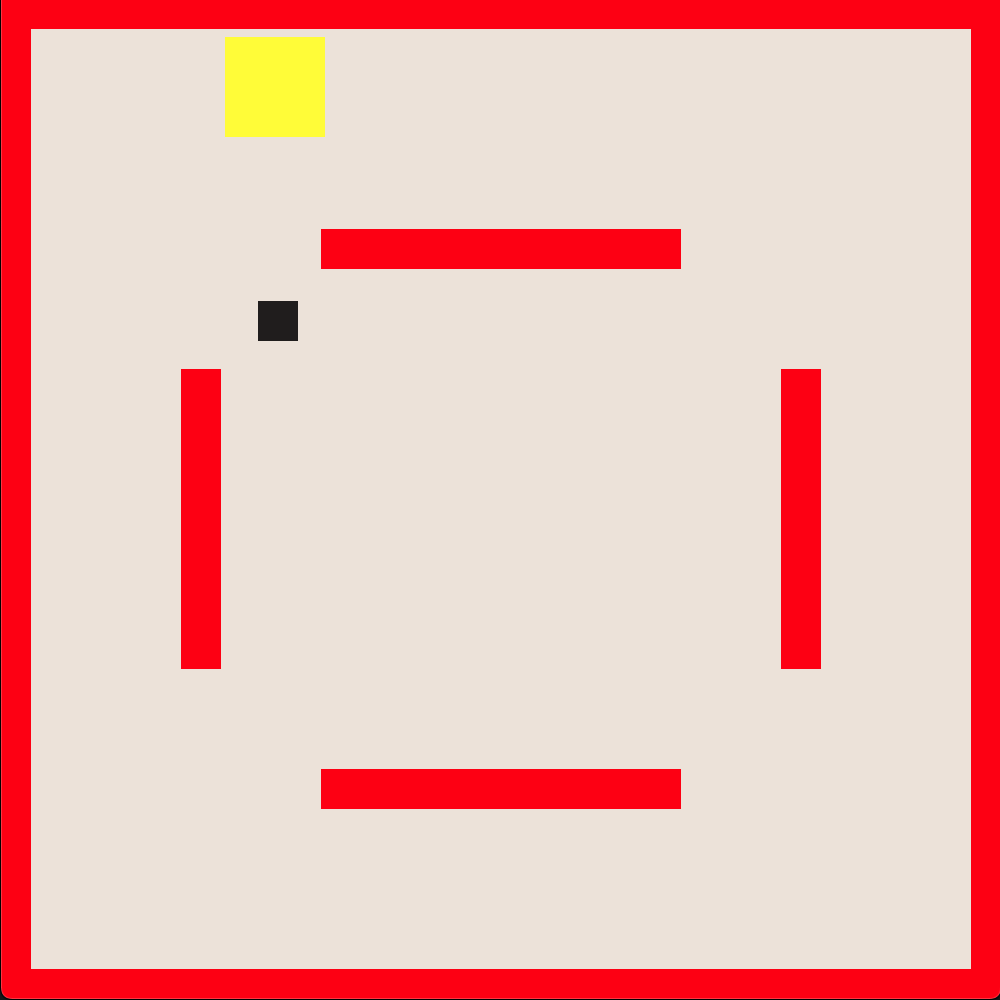}
\endminipage \hfill
\minipage{0.24\linewidth}
    \includegraphics[width=\linewidth]{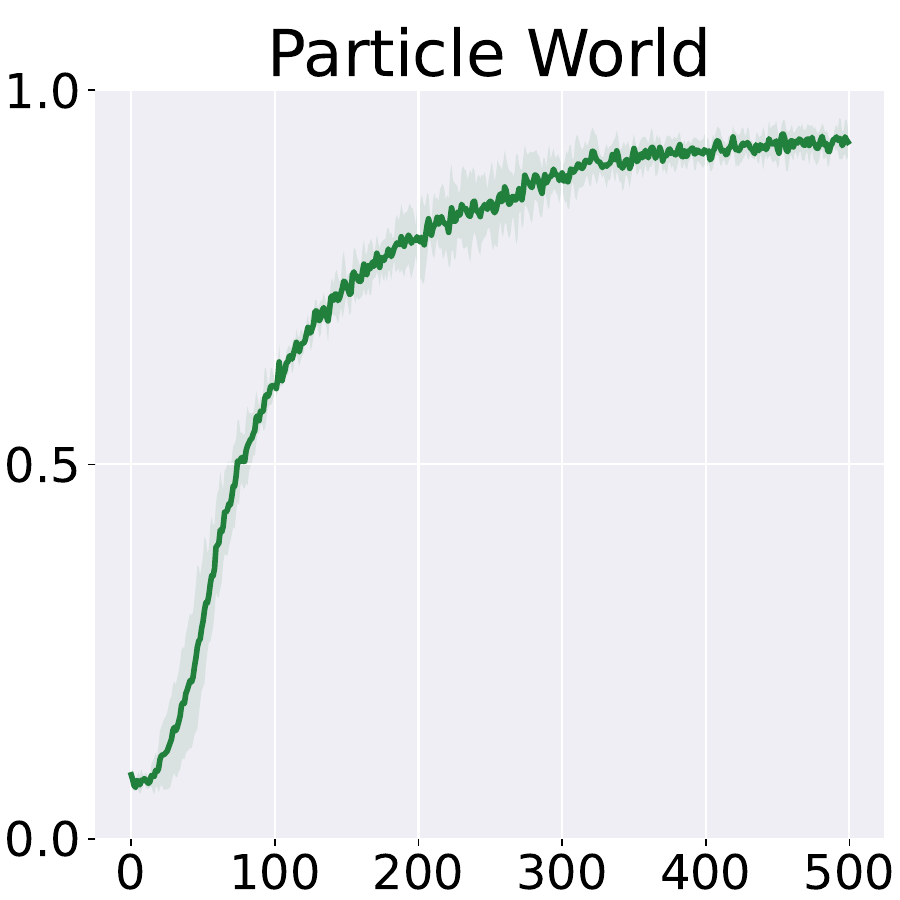}
\endminipage \hfill
\minipage{0.22\linewidth}
    \includegraphics[width=\linewidth]{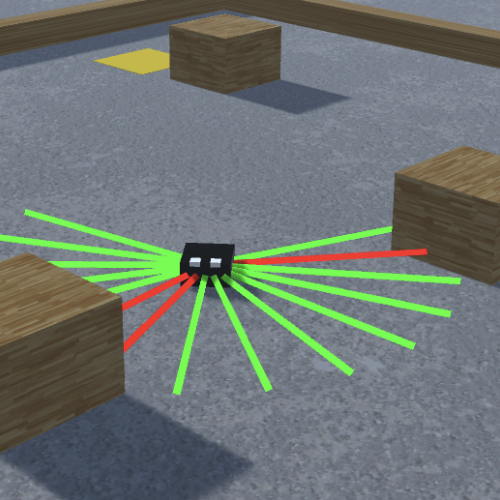}
\endminipage  \hfill
\minipage{0.24\linewidth}
    \includegraphics[width=\linewidth]{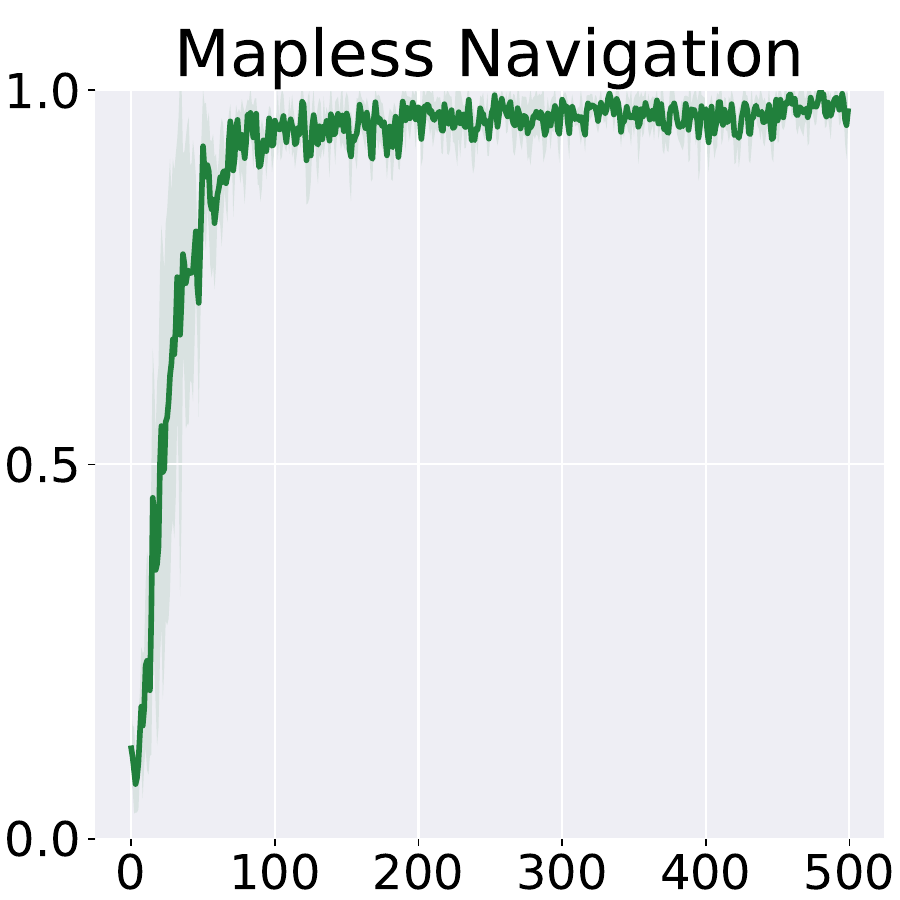}
\endminipage
\caption{The environments analyzed in our evaluation: \gridenv{} (left) and \mapless{} (right). The plot displays the number of episodes on the x-axis and the corresponding success rate on the y-axis.} 
\label{fig:motivation:training}
\end{figure}

\section{Motivation, Benchmarks, and Problem Formulation}
\label{sec:motivation}

\noindent \textbf{Benchmarks.} In our evaluation, we focused on two popular DRL benchmarks: (i) \gridenv{}, in which an agent moves in a simple two-dimensional grid trying to reach a target position while avoiding collisions with obstacles; and (ii) \mapless{}, a real-world robotic navigation task, in which a robotic agent navigates in an unknown arena by relying only on local sensors. Both benchmarks are extensively studied in the context of safe DRL~\citep{marchesini2022exploring, AmCoYeMaHaFaKa23,corsi2024analyzing} given their straightforward safety requirements (e.g., collision avoidance). For a more detailed description of the environment and training setup, see Appendix~\ref{sec:app:training}. 

\noindent \textbf{Experimental Setting.}
We extensively trained more than $250$ agents on each of these tasks, with the state-of-the-art PPO algorithm~\citep{ShWoDh17} for $500$ episodes. All agents shared the same architecture and differed solely in the random seed used to generate their initial parameters. As can be seen in Fig.~\ref{fig:motivation:training}, in both benchmarks, the trained policies reached an average \textit{success rate} (i.e., number of successful trajectories) of over $90\%$. Next, we selected per each benchmark, the five best-performing models and analyzed their performance from a safety perspective, as summarized in Tab.~\ref{tab:motivations:training}. 

\noindent \textbf{Motivation.}
All five models attained an average success rate of approximately $95\%$, and also a (seemingly) safe decision-making policy: in $100$ randomly generated trials, not a single collision was recorded. However, when analyzed through the lens of formal verification, we identified that \emph{all} the selected models had input configurations in which the policies can indeed behave unsafely and collide with a wall (see the rightmost column of Tab.~\ref{tab:motivations:training}). We believe this further motivates our work --- although the models were extensively trained with a state-of-the-art algorithm to solve a (relatively) simple task, and although they seemed to behave safely in empirical evaluation, this was indeed not the case, as formal methods were able to uncover input configurations in which they fail miserably.

\subsection{Model Selection via Verification}

The well-known susceptibility of DNNs in general, and DRL agents in particular, to adversarial inputs renders it unlikely to find models that \emph{always} behave safely across the whole input domain, even when trained for relatively simple tasks~\citep{CaKoDaKoKaAmRe22, AmCoYeMaHaFaKa23, corsi2024analyzing}. This phenomenon has also been confirmed by our evaluation reported in Tab.~\ref{tab:motivations:training}: even when formally analyzing hundreds of trained models with near-perfect empirical performance, not a single model was found to be safe throughout the entire input domain. Moreover, as discussed in Sec.~\ref{sec:preliminaries}, formal verification algorithms can detect (offline) whether a DNN is unsafe or not, however, once a DNN is deemed unsafe, it is not clear what practitioners should do. 
We believe that both aforementioned limitations further motivate the need for shielding.

\begin{table}[t]
\centering
\begin{tabular}{|c||c|c||c|}
\hline
Seed & Empirical Success (\%) & Empirical Collisions (\%) & Verification Output \\
\hline
12 & $95.6$ & $0.0$ & \sat  \\ 
 66 & $97.3$ & $0.0$ & \sat  \\ 
239 & $91.3$ & $0.0$ & \sat  \\ 
251 & $95.3$ & $0.0$ & \sat  \\ 
258 & $94.2$ & $0.0$ & \sat  \\ 
\hline
\end{tabular}
\caption{Results of the formal analysis for the top five models trained on the \gridenv{} environment. A \sat{} verification output indicates the existence of unsafe behaviors.}
\label{tab:motivations:training}
\vspace{-10pt}
\end{table}

\begin{table}[b]
\centering
\begin{tabular}{|c||c||c|c|c|}
\hline
Seed & w/o Shield Collisions (\%) & w/ Shield Collisions (\%) & Interventions (\%)  & Overhead \\
\hline
12 & $0.33$ & $0.0$ & $9.6$  & $40.0\times$ \\
 66 & $0.21$ & $0.0$ & $5.6$  & $32.5\times$ \\
 239 & $0.27$ & $0.0$ & $5.3$  & $36.3\times$ \\
251 & $0.41$ & $0.0$ & $11.0$ & $31.1\times$ \\
258 & $0.62$ & $0.0$ & $10.9$ & $35.5\times$ \\
\hline
\end{tabular}
\caption{Overhead due to standard shielding; the first two columns demonstrate how the shield can prevent collisions while introducing a significant overhead, even though fewer than $9\%$ of the actions are overridden on average across seeds. Data is collected from the \gridenv{} environment.}
\label{tab:motivations:shield-overhead}
\end{table}

\subsection{Shielding Policies in Particle Domain}
Shield synthesis is the logic-based procedure of generating a shield corresponding to an LTL specification $\varphi$, as explained briefly in Sec.~\ref{sec:preliminaries}. 
This process builds upon
encoding Boolean predicates that represent the various input variables, and hence, shield synthesis is
usually geared towards a finite state space.
However, in most DRL use cases, including the ones covered here, there is an infinite input domain, for example, the continuous input space of \gridenv{}. Still, it has very recently been proven that the task of synthesizing shields for such cases (formally known as LTL modulo first-order theories), is decidable~\citep{RoCe23a,RoCe24b} for various cases pertaining to the temporal logic encodings of $\varphi$.  Building upon these results, \cite{RoCe24} presents a novel technique that can be leveraged to synthesize shields in such scenarios~\citep{RoAmCoSaKa24}, which we use in this paper. 
We also note that the shield synthesis procedure can be expedited in various cases. For example, many specifications of interest are in the form $\varphi \coloneqq \Always \varphi'$, where $\varphi'$ is free of temporal operators. In such cases, the synthesis process can be significantly optimized and computed with alternative runtime enforcement methods~\citep{CasLaf1999, FalFerMou12}. Note that the shield both checks the correctness of the original action and provides the corrected action, when necessary. 

Tab.~\ref{tab:motivations:shield-overhead} summarizes the safety of five trained DRL policies, with and without shielding. Although the shield's soundness indeed guarantees absolute safety (see the third column indicating no collisions), our results demonstrate the main limitations of current shielding approaches: the shield is activated \emph{online} in every time-step throughout the \emph{whole} input domain, resulting in a significant overhead during deployment. From our preliminary analysis, we found that in most cases the shield does not override the original action (i.e., no interventions; see fourth column), demonstrating the safety of the original behavior in the majority of cases. However, even in such situations, the overhead remains due to the constant shield execution. In the following section, we propose an approach that takes advantage of this and drastically reduces the number of calls to the shield while still guaranteeing safety throughout the entire input domain.

\section{Verification-Guided Shielding}
\label{sec:method}

Next, building on the previous findings, we present our novel
\textit{verification-driven shielding} approach. We devised this
method using recent advances in formal verification, shielding, and symbolic representation. Our approach incorporates five steps: (1) domain splitting, after which we perform (2) safe-region verification, and (3) clustering. Subsequently, we (4) encode a symbolic representation of the input domain, which finally allows (5) shielding the agent only on potentially unsafe regions.

\paragraph{(1) Domain Splitting.}
In the first step, our method employs an off-the-shelf verification algorithm to identify all the regions of the state space in which a given DRL agent behaves correctly. This process mainly relies on the concept of \textit{All-DNN-Verification}~\citep{marzari2023enumerating}. In essence, this includes pruning the input region in search of all regions in which the trained agent is provably safe, with respect to a set of given requirements. More formally, we search for regions in which the negated (safety) property is \unsat{}. An exact solution to this problem would provide the complement of the regions where the agent potentially requires a shield. 

However, given that this problem is \#P-hard~\citep{marzari2023enumerating}, the authors proposed \texttt{$\epsilon$-ProVe}, an algorithm that computes an underapproximation of these safe regions. In more detail, \texttt{$\epsilon$-ProVe} divides the input domain into regions, effectively generating a search tree where each node represents a partition of the input domain; these regions are then analyzed using a sampling approach that provides an estimated probability that the region is safe. The algorithm iteratively splits regions into subregions, until it cannot find any counterexamples (i.e., \unsat{} assignments to the negated property), in which case the region is approximated as safe.
Otherwise, the algorithm heuristically decides whether to declare the entire region as unsafe or continue with the splitting procedure. For a detailed description of \texttt{$\epsilon$-ProVe}, and a discussion regarding the probabilistic guarantees provided, we refer the reader to~\citep{marzari2023enumerating}.

\paragraph{(2) Formal Verification of Safe Regions.}
Subsequently, we are left with a division of the input domain into regions, with each region approximated as either safe or unsafe. Although \texttt{$\epsilon$-ProVe} typically provides tight results with high confidence, the approximated nature of the approach is not enough to guarantee \emph{absolute} correctness, which is the subject matter of this work. To address this gap, we complement the approximated regions by formally verifying the regions previously approximated as safe. Toward this end, we employ \texttt{Marabou}, a sound and complete verification tool~\citep{KaHuIbJuLaLiShThWuZeDiKoBa19,WuIsZeTaDaKoReAmJuBaHuLaWuZhKoKaBa24}, which is used to formally certify the safety of the agent only in the regions that are previously approximated as safe. 
In this second, fined-tuned verification procedure, if a counterexample (\sat{}) is found in a (mistakenly approximated) safe region, we reclassify the region as unsafe. On the other hand, regions that have already been found to violate the property (\texttt{$\epsilon$-ProVe} returned \sat{} in the first step), are left untouched, as a valid counterexample was already found. A pseudocode describing this procedure can be found in Appendix~\ref{sec:app:verification-code}.



\begin{figure}[h]
\centering
    \includegraphics[width=\linewidth]{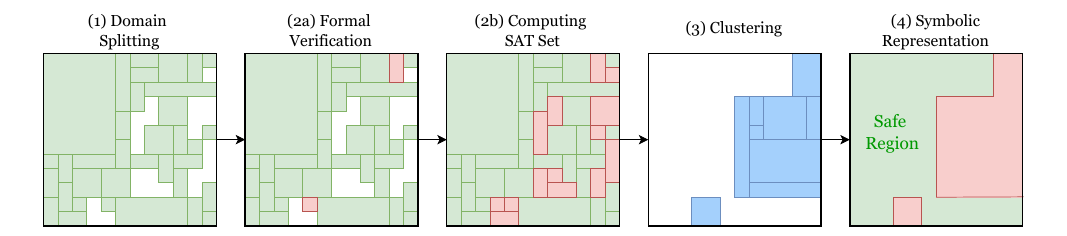}
\caption{
An overview of \textit{verification-guided shielding}. In step (1) we employ \texttt{$\epsilon$-ProVe} to split the input domain into approximated safe (green) and unsafe regions; (2a) these can be further validated with a formal verification tool, (2b) which complements the set of approximated unsafe regions with the ones formally found as such. (3) To reduce the cardinality of this set we employ a clustering algorithm and (4) further simplify the encoded results by using symbolic representation.
}
\label{fig:method:algorithm}
\end{figure}

\paragraph{(3) Clustering.} 
After these first two steps, we are left with a sound division of the input space into regions in which the agent is provably safe and regions in which there is at least one input configuration that causes the agent to behave unsafely.
Next, we would like to apply our synthesized shield solely on these potentially unsafe regions. However, this presents a new challenge as the set $\mathcal{S}$ of unsafe regions includes, in practice, a large number of compact regions (e.g., we observed an average cardinality of $\approx60,000$ in \gridenv). While this does not compromise the correctness of our strategy, it raises another problem --- checking whether the current input belongs to the set $\mathcal{S}$ introduces significant overhead, potentially mitigating the benefits of our approach. To address this limitation, we employ an additional step in which we cluster the set of unsafe regions and reduce their overall cardinality. Specifically, we employ \textit{agglomerative clustering}~\citep{ackermann2014analysis}, to concatenate unsafe regions and produce an overapproximation of the unsafe regions. As we later demonstrate, this \emph{significantly} reduces the number of unsafe regions and, consequently, the overhead for checking whether the current state belongs to $\mathcal{S}$. 
It is important to emphasize that, although the clustering step has the potential to overapproximate safe regions as unsafe (see Fig.~\ref{fig:method:algorithm}), it does not compromise the overall soundness of our approach. At most, the shield may be activated more than strictly necessary.

\paragraph{(4) Symbolic Representation.} 
Next, we make use of symbolic representation~\citep{HoGoSelKay07}, i.e., an encoding of all the states in order to obtain a succinct formula for all unsafe regions. Due to our focus on fully observable input domains, we can use propositional logic formulas to symbolically encode the regions of interest (in our case, unsafe regions). Furthermore, this formula can be reduced to a succinct equivalent formula, e.g., by using off-the-shelf solvers~\citep{DeBj08,BaTi18}. 
This, in turn, could potentially reduce the overhead of checking online whether the agent is in an unsafe region. We elaborate further on symbolic representation in Appendix~\ref{sec:app:symbolic}.
%

\paragraph{(5) Shield Synthesis and Execution.}
Finally, we are left with a set of (relatively) few approximated unsafe regions. First, we can synthesize a shield that, when activated, guarantees safety with respect to the safety property $\psi$. Next, we can couple the shield with the agent, and at each time-step: (i) efficiently identify if the current input belongs to the potentially unsafe regions, and if so, (ii) temporarily activate the synthesized shield and guarantee the safety, as previously described in Sec.~\ref{sec:motivation}. In the remaining (provably) safe regions, the shield can remain inactive while preserving the formal guarantees, as we already verified that any original decision that the agent makes is safe.

\section{Empirical Evaluation}
\label{sec:evaluation}

Our experimental evaluations comprise two components: the \textit{offline} procedures for generating the shield and identifying safe regions, and the \textit{online} execution of the system, where the goal is to minimize the computational overhead resulting from invoking the shield. 

\noindent \textbf{Experimental Setup.} 
The offline evaluation was conducted on a distributed cluster with 160 CPUs and 448GB RAM. For each verification query, we employ 1 CPU, 1GB RAM, and a runtime limit of three hours. 
We also note that in the case of the verification queries, Marabou internally used the \emph{Guorobi} LP solver~\citep{AnAgKu17} as a backend engine.
Data related to the overhead was collected on a commercial laptop to align with the limited hardware typically used to operate autonomous robotic systems.

\noindent \textbf{Offline Procedures.} Tab.~\ref{tab:evaluation:offline} summarizes the time measured for each of the offline steps with respect to the \gridenv{} benchmark. The most time-consuming component is the formal verification (step $2$), taking an average of over two hours; this is not surprising, as in this step the verifier is required to solve many NP-complete problems, per each policy. However, we believe this is a reasonable price as (i) this is an offline step that is executed once; and (ii) this is the step that provides the formal guarantees. On the other hand, \texttt{$\epsilon$-ProVe} (i.e., the splitting procedure in step $1$) runs significantly faster but provides only probabilistic assurances. Both stages are complementary and represent a simulated annealing-like optimization~\citep{KiGeVe83}: at first, we approximate and reduce the number of regions on which we can rule out correct behavior, and then, we run a more expensive, formal verification procedure that fine-tunes the remaining regions. The column representing the clustering demonstrates a high variance among the different policies, as there can be significant differences in the number of unsafe regions identified in the previous steps. Still, it is worth noting that even in the worst-case measurement, the time required by this procedure is negligible when compared to the formal verification step. Finally, the table reports the results of the shield synthesis and the formula reduction steps, which were not particularly time-demanding (i.e., step 4). These results also align with the hypothesis raised in the previous sections, i.e., that the heavy computational cost of shielding is not related to the offline synthesis time, but rather to the cost of invoking the shield online, before each decision. 

\begin{table}[t]
\centering
\begin{tabular}{|c||c|c|c|c|c|}
\hline
Seed & Splitting (s) & Verification (hr) & Clustering (s) & Synthesis (s) & Reduction (s) \\
\hline
12 & $251.7$ & $1.44$ & $ 12.28$ & \multirow{5}{*}{$2.69$}  & $22.48$ \\
66 & $239.1$ & $2.01$ & $ 14.02$ &  & $29.12$ \\
239 & $458.0$ & $2.17$ & $120.84$ &  & $38.67$ \\
251 & $451.2$ & $ 2.26$ & $139.30$ &  & $38.93$ \\
258 & $485.4$ & $2.23$ & $248.55$ & & $43.62$ \\
\hline
\end{tabular}
\caption{\gridenv{}: the overall time required for the offline components of our approach.}
\label{tab:evaluation:offline}
\vspace{-10pt}
\end{table}

\begin{table}[b!]
\centering
\begin{tabular}{|c||cc||cc||c|}
\hline
\textbf{Seed} & \multicolumn{2}{c||}{\textbf{Full Shield}} & \multicolumn{2}{c||}{\textbf{Verification-Guided Shield}}  & \textbf{Gain (\%)} \\ \cline{2-5}
     & \multicolumn{1}{c|}{Active Time (\%)} & Overhead & \multicolumn{1}{c|}{Active Time (\%)} & Overhead & \\ \hline
 12  & \multicolumn{1}{c|}{$100$} & $40.0\times$   & \multicolumn{1}{c|}{$28.6$} & $14.1\times$   & $64.8$  \\ 
66   & \multicolumn{1}{c|}{$100$} & $32.5\times$   & \multicolumn{1}{c|}{$32.4$} & $13.1\times$   & $59.7$  \\
239  & \multicolumn{1}{c|}{$100$} & $36.3\times$   & \multicolumn{1}{c|}{$44.5$} & $21.5\times$   & $40.7$  \\
251  & \multicolumn{1}{c|}{$100$} & $31.1\times$   & \multicolumn{1}{c|}{$37.6$} & $13.2\times$   & $57.6$  \\
258  & \multicolumn{1}{c|}{$100$} & $35.5\times$   & \multicolumn{1}{c|}{$33.8$} & $13.9\times$   & $60.1$  \\
\hline
104  & \multicolumn{1}{c|}{$100$} & $4.8\times$   & \multicolumn{1}{c|}{$61.7$} & $3.6\times$   & $25.1$  \\
225  & \multicolumn{1}{c|}{$100$} & $4.4\times$   & \multicolumn{1}{c|}{$53.1$} & $3.5\times$   & $20.5$  \\
239  & \multicolumn{1}{c|}{$100$} & $4.5\times$   & \multicolumn{1}{c|}{$2.1$}  & $1.8\times$   & $60.0$  \\
243  & \multicolumn{1}{c|}{$100$} & $4.5\times$   & \multicolumn{1}{c|}{$1.3$}  & $1.6\times$   & $71.1$  \\ 
310  & \multicolumn{1}{c|}{$100$} & $4.6\times$   & \multicolumn{1}{c|}{$3.4$}  & $1.5\times$   & $67.4$  \\
\hline
\end{tabular}
\caption{Final results; the first block presents the results for the \gridenv{} benchmark and the second block represents the results for the \mapless{} benchmark.}
\label{tab:evaluation:main}
\end{table}

\noindent \textbf{Online Invocation.} 
Our main results are presented in Tab.~\ref{tab:evaluation:main}. Specifically, we compare the overhead introduced by the shield in two cases: the classic fully-activated shielding approach and our verification-guided approach. The first half of the table reports the analysis on the \gridenv{} environment, while the second one reports the results for \mapless{}. In general, both benchmarks demonstrate the merits of our approach in reducing significant overhead, confirming the general environment nature of our methodology. In more detail, per each environment and seed, we compare the portion of time in which the shield was activated during execution. This value is computed by normalizing the number of shield invocations by the total number of actions. While in the first column, the (full) shield is trivially always active, we observe a drastic reduction when our approach has been applied, especially in some \mapless{} seeds. We also note that there is not necessarily a direct correlation between the size of the unsafe regions and the number of interventions, as the interventions were measured with respect to stochastic executions. Finally, we report the average computational time overhead as a relative value compared to an episode with the shield deactivated, i.e., decisions made without invoking any external components. Not surprisingly, our results show a clear correlation between the \textit{active time} and the overhead introduced, further motivating this work.  In the last column, we summarize the time gain provided by our method with respect to invoking the shield at every time-step. 
Note that the gain is not always proportional to the \textit{active time}; the resulting value also depends on the actual number of steps required to complete a single episode, i.e., the absolute number of calls to the shield.

\textbf{Note.} Per each benchmark, we encoded $1,000$ LTL formulas for our shield. We emphasize that this number does not affect the relative gain of our \textit{verification-guided shield}, but only affects the \emph{absolute} value of the overhead compared to a single forward propagation. We also emphasize that the formal verification was skipped in \mapless{}, in order to expedite the procedure, and hence we relied only on the probabilistic guarantees afforded by \texttt{$\epsilon$-ProVe}.
In addition, we note that our evaluation included the first three steps, while excluding the fourth step, i.e., symbolic representation, as this step ran slightly slower than when using the complete set of input regions. 
Still, we executed this step and reported the overall time that it took, while successfully demonstrating that symbolic representations can be encoded for environments including thousands of states.
Since the efficiency of this step highly depends on the task in question and the underlying SMT solver, improving symbolic representations is beyond the scope of this work. 

\textbf{Limitations.} Although these results are encouraging, it is important to acknowledge certain limitations of our approach, mainly inherited from the backend shielding and verification techniques.  First, our approach requires a valid encoding of the required properties of interest. This, in turn, assumes access to the environment dynamics and the agent's transition model, as well as the practitioner's ability to encode the relevant properties in a logic-based form.
In addition, when relying on shielding in unsafe regions, it is important to note that although the shield guarantees adherence to the given requirements, it does not necessarily select the \emph{optimal} action, among the safe ones. 
Furthermore, there are some limitations due to the backend DNN verification tools. Specifically, our approach relies on \texttt{Marabou}, which affords only limited support to some activation functions, hence restricting its applicability to some advanced DNN architectures. 
However, we believe that these limitations can serve as a foundation for future research.

\section{Related Work}
\label{sec:related-work}

In recent years, the formal methods community has put forth a wide range of approaches and tools for formally verifying the correctness of deep learning models~\citep{TjXiTe17, LoMa17, HuKwWaWu17, WaPeWhYaJa18, GeMiDrTsChVe18, KuKaGoJuBaKo18, GoKaPaBa18, SiGePuVe19, LyKoKoWoLiDa20, KaBaDiJuKo21, WuIsZeTaDaKoReAmJuBaHuLaWuZhKoKaBa24}. In addition, there has recently been ample research on formally verifying DRL systems~\citep{FuPl18, DuJhSaTi18b, DuChSa19, SuKhSh19, VaPeWaNiSiKh22,MaAmWuDaNeRaMeDuGaShKaBa24}, in particular in the context of safety~\citep{KaBaKaSc19, AmScKa21,ElKaKaSc21} and explainability~\citep{BaKa23,BaAmCoReKa23}. 
Other work focused on enhancing DRL safety by inducing Scenario-Based Programming (SBP)~\citep{CoYeAmFaHaKa22, YeAmElHaKaMa22, YeAmElHaKaMa23}.

Classic shielding  approaches~\citep{AlBloEh18,PraKoPoBlo21,PraKoTa21} focus on properties expressed in Boolean LTL and are incompatible with systems pertaining to richer data domains. However, more recently,~\cite{WuMaDeWa19} presented the concept of (incomplete) shields for linear arithmetic. To address these limitations, the work of~\cite{RoCe23a,RoCe24} proves that, under certain conditions, LTL synthesis with data specifications is decidable via abstraction methods,
which can be used for reactive synthesis of expressive shields with e.g., numerical information~\citep{RoAmCoSaKa24}. 
Additional techniques, such as runtime enforcement and supervisory control~\citep{CasLaf1999, Schne00,LiBauWal09,FalFerMou12}, share similarities with shielding, however, such methods are incompatible with DRL and general reactive systems, but rather, focus solely on checking software invariants. 

We also note that in addition to formal verification and shielding, there exist other popular approaches for improving the safety of DRL agents. These methods are typically applied during training and rely on constrained optimization~\citep{StAcAb20, LiDiLi20, RoGiRo21}, safe exploration~\citep{SrEyHa20, YaSiJa22}, and various alternative solutions~\citep{GaFe15, AcHeTa17,ChDaSh19}. However, although popular, these techniques are heuristic in nature and do not afford any formal guarantees regarding the safety of the DRL agent in question~\citep{BrGrHa21}.
 
\section{Conclusion}
\label{sec:conclusion}

In this paper, we combine verification and shielding and propose a novel technique that leverages the advantages of both these formal approaches. Specifically, we demonstrate how to use formal methods to prune the input space and divide it into safe and (potentially) unsafe regions. While in the first case, we can safely employ the original, shield-less model, in the latter we activate the shield online, and override any potential unsafe action. We extensively evaluate our approach in multiple experiments, and demonstrate that it drastically reduces the overhead of shielding, while maintaining guaranteed safety throughout the whole input domain. 

Moving forward, our approach can be extended along various axes. Currently, we employ clustering algorithms to approximate unsafe regions, however, we plan to investigate additional strategies that attain more compact descriptions and reduce the use of shielding even further. Additionally, we plan to incorporate our approach also into the DRL training phase to iteratively robustify the agents prior to deployment. Finally, we plan to explore alternative strategies, such as deep ensembles, to further reduce the need for shield interventions. We believe this work is another step towards the reliable use of DRL in safety-critical domains.

\subsubsection*{Acknowledgments}
The work of Corsi and Fox was funded in part by the National Science Foundation (Award \#2321786).
The work of Amir and Katz was partially funded
by the European Union (ERC, VeriDeL, 101112713). Views
and opinions expressed are however those of the author(s) only
and do not necessarily reflect those of the European Union
or the European Research Council Executive Agency. Neither
the European Union nor the granting authority can be held
responsible for them.
The work of Amir was further supported by a scholarship from the Clore Israel Foundation. 
The work of Rodríguez and Sánchez was funded in part by PRODIGY Project (TED2021-132464B-I00) --- funded by MCIN/AEI/10.13039/501100011033/ and the European Union NextGenerationEU/PRTR --- by the DECO Project (PID2022-138072OB-I00) --- funded by MCIN/AEI/10.13039/501100011033 and by the ESF, as well as by a research grant from Nomadic Labs and the Tezos Foundation.

\bibliography{main}

\begin{thebibliography}{99}
\providecommand{\natexlab}[1]{#1}
\providecommand{\url}[1]{\texttt{#1}}
\expandafter\ifx\csname urlstyle\endcsname\relax
  \providecommand{\doi}[1]{doi: #1}\else
  \providecommand{\doi}{doi: \begingroup \urlstyle{rm}\Url}\fi

\bibitem[Achiam et~al.(2017)Achiam, Held, Tamar, and Abbeel]{AcHeTa17}
J.~Achiam, D.~Held, A.~Tamar, and P.~Abbeel.
\newblock {Constrained Policy Optimization}.
\newblock In \emph{Int. Conf. on Machine Learning}, pp.\  22--31. PMLR, 2017.

\bibitem[Ackermann et~al.(2014)Ackermann, Bl{\"o}mer, Kuntze, and Sohler]{ackermann2014analysis}
M.~Ackermann, J.~Bl{\"o}mer, D.~Kuntze, and C.~Sohler.
\newblock {Analysis of Agglomerative Clustering}.
\newblock \emph{Algorithmica}, 2014.

\bibitem[Alshiekh et~al.(2018)Alshiekh, Bloem, Ehlers, K{\"{o}}nighofer, Niekum, and Topcu]{AlBloEh18}
M.~Alshiekh, R.~Bloem, R.~Ehlers, B.~K{\"{o}}nighofer, S.~Niekum, and U.~Topcu.
\newblock {Safe Reinforcement Learning via Shielding}.
\newblock In \emph{Proc. of the 32nd AAAI Conference on Artificial Intelligence}, pp.\  2669--2678, 2018.

\bibitem[Amir et~al.(2021{\natexlab{a}})Amir, Schapira, and Katz]{AmScKa21}
G.~Amir, M.~Schapira, and G.~Katz.
\newblock {Towards Scalable Verification of Deep Reinforcement Learning}.
\newblock In \emph{Proc. 21st Int. Conf. on Formal Methods in Computer-Aided Design (FMCAD)}, pp.\  193--203, 2021{\natexlab{a}}.

\bibitem[Amir et~al.(2021{\natexlab{b}})Amir, Wu, Barrett, and Katz]{AmWuBaKa21}
G.~Amir, H.~Wu, C.~Barrett, and G.~Katz.
\newblock {An SMT-Based Approach for Verifying Binarized Neural Networks}.
\newblock In \emph{Proc. 27th Int. Conf. on Tools and Algorithms for the Construction and Analysis of Systems (TACAS)}, pp.\  203--222, 2021{\natexlab{b}}.

\bibitem[Amir et~al.(2022)Amir, Zelazny, Katz, and Schapira]{AmZeKaSc22}
G.~Amir, T.~Zelazny, G.~Katz, and M.~Schapira.
\newblock {Verification-Aided Deep Ensemble Selection}.
\newblock In \emph{Proc. 22nd Int. Conf. on Formal Methods in Computer-Aided Design (FMCAD)}, pp.\  27--37, 2022.

\bibitem[Amir et~al.(2023{\natexlab{a}})Amir, Corsi, Yerushalmi, Marzari, Harel, Farinelli, and Katz]{AmCoYeMaHaFaKa23}
G.~Amir, D.~Corsi, R.~Yerushalmi, L.~Marzari, D.~Harel, A.~Farinelli, and G.~Katz.
\newblock {Verifying Learning-Based Robotic Navigation Systems}.
\newblock In \emph{Proc. 29th Int. Conf. on Tools and Algorithms for the Construction and Analysis of Systems (TACAS)}, pp.\  607--627, 2023{\natexlab{a}}.

\bibitem[Amir et~al.(2023{\natexlab{b}})Amir, Freund, Katz, Mandelbaum, and Refaeli]{AmFrKaMaRe23}
G.~Amir, Z.~Freund, G.~Katz, E.~Mandelbaum, and I.~Refaeli.
\newblock {veriFIRE: Verifying an Industrial, Learning-Based Wildfire Detection System}.
\newblock In \emph{Proc. 25th Int. Symposium on Formal Methods (FM)}, pp.\  648--656, 2023{\natexlab{b}}.

\bibitem[Amir et~al.(2023{\natexlab{c}})Amir, Maayan, Zelazny, Katz, and Schapira]{AmMaZeKaSc23}
G.~Amir, O.~Maayan, T.~Zelazny, G.~Katz, and M.~Schapira.
\newblock {Verifying Generalization in Deep Learning}.
\newblock In \emph{Proc. 35th Int. Conf. on Computer Aided Verification (CAV)}, pp.\  438--455, 2023{\natexlab{c}}.

\bibitem[Anand et~al.(2017)Anand, Aggarwal, and Kumar]{AnAgKu17}
R.~Anand, D.~Aggarwal, and V.~Kumar.
\newblock {A Comparative Analysis of Optimization Solvers}.
\newblock \emph{Journal of Statistics and Management Systems}, 20\penalty0 (4):\penalty0 623--635, 2017.

\bibitem[Aractingi et~al.(2023)Aractingi, L{\'e}ziart, Flayols, Perez, Silander, and Sou{\`e}res]{aractingi2023controlling}
M.~Aractingi, P.~L{\'e}ziart, T.~Flayols, J.~Perez, T.~Silander, and P.~Sou{\`e}res.
\newblock {Controlling the Solo12 Quadruped Robot with Deep Reinforcement Learning}, 2023.

\bibitem[Barrett \& Tinelli(2018)Barrett and Tinelli]{BaTi18}
C.~Barrett and C.~Tinelli.
\newblock \emph{{Satisfiability Modulo Theories}}.
\newblock Springer, 2018.

\bibitem[Bassan \& Katz(2023)Bassan and Katz]{BaKa23}
S.~Bassan and G.~Katz.
\newblock {Towards Formal Approximated Minimal Explanations of Neural Networks}.
\newblock In \emph{Proc. 29th Int. Conf. on Tools and Algorithms for the Construction and Analysis of Systems (TACAS)}, pp.\  187--207, 2023.

\bibitem[Bassan et~al.(2023)Bassan, Amir, Corsi, Refaeli, and Katz]{BaAmCoReKa23}
S.~Bassan, G.~Amir, D.~Corsi, I.~Refaeli, and G.~Katz.
\newblock {Formally Explaining Neural Networks within Reactive Systems}.
\newblock In \emph{Proc. 23rd Int. Conf. on Formal Methods in Computer-Aided Design (FMCAD)}, pp.\  10--22, 2023.

\bibitem[Bloem et~al.(2015)Bloem, K{\"{o}}nighofer, K{\"{o}}nighofer, and Wang]{BlKoKoWa15}
R.~Bloem, B.~K{\"{o}}nighofer, R.~K{\"{o}}nighofer, and C.~Wang.
\newblock {Shield Synthesis: - Runtime Enforcement for Reactive Systems}.
\newblock In \emph{Proc. of the 21st Int. Conf. in Tools and Algorithms for the Construction and Analysis of Systems, (TACAS)}, volume 9035, pp.\  533--548, 2015.

\bibitem[Brunke et~al.(2021)Brunke, Greeff, Hall, Yuan, Zhou, Panerati, and Schoellig]{BrGrHa21}
L.~Brunke, M.~Greeff, A.~Hall, Z.~Yuan, S.~Zhou, J.~Panerati, and A.~Schoellig.
\newblock {Safe Learning in Robotics: From Learning-Based Control to Safe Reinforcement Learning}.
\newblock \emph{Annual Review of Control, Robotics, and Autonomous Systems}, 5, 2021.

\bibitem[Casadio et~al.(2022)Casadio, Komendantskaya, Daggitt, Kokke, Katz, Amir, and Refaeli]{CaKoDaKoKaAmRe22}
M.~Casadio, E.~Komendantskaya, M.~Daggitt, W.~Kokke, G.~Katz, G.~Amir, and I.~Refaeli.
\newblock {Neural Network Robustness as a Verification Property: A Principled Case Study}.
\newblock In \emph{Proc. 34th Int. Conf. on Computer Aided Verification (CAV)}, pp.\  219--231, 2022.

\bibitem[Cassandras \& Lafortune(1999)Cassandras and Lafortune]{CasLaf1999}
C.~Cassandras and S.~Lafortune.
\newblock \emph{{Introduction to Discrete Event Systems}}, volume~11 of \emph{The Kluwer International Series on Discrete Event Dynamic Systems}.
\newblock Springer, 1999.

\bibitem[Corsi et~al.(2021)Corsi, Marchesini, and Farinelli]{CoMaFa21}
D.~Corsi, E.~Marchesini, and A.~Farinelli.
\newblock {Formal Verification of Neural Networks for Safety-Critical Tasks in Deep Reinforcement Learning}.
\newblock In \emph{Proc. 37th Conf. on Uncertainty in Artificial Intelligence (UAI)}, 2021.

\bibitem[Corsi et~al.(2022)Corsi, Yerushalmi, Amir, Farinelli, Harel, and Katz]{CoYeAmFaHaKa22}
D.~Corsi, R.~Yerushalmi, G.~Amir, A.~Farinelli, D.~Harel, and G.~Katz.
\newblock {Constrained Reinforcement Learning for Robotics via Scenario-Based Programming}, 2022.
\newblock Technical Report. \url{https://arxiv.org/abs/2206.09603}.

\bibitem[Corsi et~al.(2024)Corsi, Amir, Katz, and Farinelli]{corsi2024analyzing}
D.~Corsi, G.~Amir, G.~Katz, and A.~Farinelli.
\newblock {Analyzing Adversarial Inputs in Deep Reinforcement Learning}, 2024.
\newblock Technical Report. \url{https://arxiv.org/abs/2402.05284}.

\bibitem[De~Moura \& Bj{\o}rner(2008)De~Moura and Bj{\o}rner]{DeBj08}
L.~De~Moura and N.~Bj{\o}rner.
\newblock {Z3: An Efficient SMT Solver}.
\newblock In \emph{Proc. 14th Int. Conf. on Tools and Algorithms for the Construction and Analysis of Systems (TACAS)}, pp.\  337--340, 2008.

\bibitem[Dutta et~al.(2018)Dutta, Jha, Sankaranarayanan, and Tiwari]{DuJhSaTi18b}
S.~Dutta, S.~Jha, S.~Sankaranarayanan, and A.~Tiwari.
\newblock {Learning and Verification of Feedback Control Systems using Feedforward Neural Networks}.
\newblock \emph{IFAC-PapersOnLine}, 51\penalty0 (16):\penalty0 151--156, 2018.

\bibitem[Dutta et~al.(2019)Dutta, Chen, and Sankaranarayanan]{DuChSa19}
S.~Dutta, X.~Chen, and S.~Sankaranarayanan.
\newblock {Reachability Analysis for Neural Feedback Systems using Regressive Polynomial Rule Inference}.
\newblock In \emph{Proc. 22nd ACM Int. Conf. on Hybrid Systems: Computation and Control (HSCC)}, pp.\  157--168, 2019.

\bibitem[Elboher et~al.(2020)Elboher, Gottschlich, and Katz]{ElGoKa20}
Y.~Elboher, J.~Gottschlich, and G.~Katz.
\newblock {An Abstraction-Based Framework for Neural Network Verification}.
\newblock In \emph{Proc. 32nd Int. Conf. on Computer Aided Verification (CAV)}, pp.\  43--65, 2020.

\bibitem[Elboher et~al.(2022)Elboher, Cohen, and Katz]{ElCoKa22}
Y.~Elboher, E.~Cohen, and G~Katz.
\newblock {Neural Network Verification using Residual Reasoning}.
\newblock In \emph{Proc. 20th Int. Conf. on Software Engineering and Formal Methods (SEFM)}, pp.\  173--189, 2022.

\bibitem[Eliyahu et~al.(2021)Eliyahu, Kazak, Katz, and Schapira]{ElKaKaSc21}
T.~Eliyahu, Y.~Kazak, G.~Katz, and M.~Schapira.
\newblock {Verifying Learning-Augmented Systems}.
\newblock In \emph{Proc. Conf. of the ACM Special Interest Group on Data Communication on the Applications, Technologies, Architectures, and Protocols for Computer Communication (SIGCOMM)}, pp.\  305--318, 2021.

\bibitem[Falcone et~al.(2012)Falcone, Fernandez, and Mounier]{FalFerMou12}
Y.~Falcone, J.~Fernandez, and L.~Mounier.
\newblock {What can you Verify and Enforce at Runtime?}
\newblock \emph{Int. Journal on Software Tools for Technology Transfer}, pp.\  349--382, 2012.

\bibitem[Ferhat \& Yildirim-Yayilgan(2020)Ferhat and Yildirim-Yayilgan]{FeYi20}
O.~Ferhat and S.~Yildirim-Yayilgan.
\newblock {Deep Neural Network Based Malicious Network Activity Detection Under Adversarial Machine Learning Attacks}.
\newblock In \emph{Proc. 3rd Int. Conf. on Intelligent Technologies and Applications (INTAP)}, pp.\  280--291, 2020.

\bibitem[Fulton \& Platzer(2018)Fulton and Platzer]{FuPl18}
N.~Fulton and A.~Platzer.
\newblock {Safe Reinforcement Learning via Formal Methods: Toward Safe Control through Proof and Learning}.
\newblock In \emph{Proc. 32nd AAAI Conf. on Artificial Intelligence (AAAI)}, 2018.

\bibitem[Garc{\i}a \& Fern{\'a}ndez(2015)Garc{\i}a and Fern{\'a}ndez]{GaFe15}
J.~Garc{\i}a and F.~Fern{\'a}ndez.
\newblock {A Comprehensive Survey on Safe Reinforcement Learning}.
\newblock \emph{Journal of Machine Learning Research}, 16\penalty0 (1):\penalty0 1437--1480, 2015.

\bibitem[Gehr et~al.(2018)Gehr, Mirman, Drachsler-Cohen, Tsankov, Chaudhuri, and Vechev]{GeMiDrTsChVe18}
T.~Gehr, M.~Mirman, D.~Drachsler-Cohen, E.~Tsankov, S.~Chaudhuri, and M.~Vechev.
\newblock {AI2: Safety and Robustness Certification of Neural Networks with Abstract Interpretation}.
\newblock In \emph{Proc. 39th IEEE Symposium on Security and Privacy (S\&P)}, 2018.

\bibitem[Gongye et~al.(2020)Gongye, Li, Zhang, Sabbagh, Yuan, Lin, Wahl, and Fei]{GoLiZhSaYuLiWaFe20}
C.~Gongye, H.~Li, X.~Zhang, M.~Sabbagh, G.~Yuan, X.~Lin, T.~Wahl, and Y.~Fei.
\newblock {New Passive and Active Attacks on Deep Neural Networks in Medical Applications}.
\newblock In \emph{Proceedings of the 39th International Conference on Computer-Aided Design}, pp.\  1--9, 2020.

\bibitem[Gopinath et~al.(2018)Gopinath, Katz, P\v{a}s\v{a}reanu, and Barrett]{GoKaPaBa18}
D.~Gopinath, G.~Katz, C.~P\v{a}s\v{a}reanu, and C.~Barrett.
\newblock {DeepSafe: A Data-driven Approach for Assessing Robustness of Neural Networks}.
\newblock In \emph{Proc. 16th. Int. Symposium on Automated Technology for Verification and Analysis (ATVA)}, pp.\  3--19, 2018.

\bibitem[Hoffmann et~al.(2007)Hoffmann, Gomes, Selman, and Kautz]{HoGoSelKay07}
J.~Hoffmann, C.~Gomes, B.~Selman, and H.~Kautz.
\newblock {SAT Encodings of State-Space Reachability Problems in Numeric Domains}.
\newblock In \emph{Proc. of the 20th Int. Joint Conf. on Artificial Intelligence (IJCAI)}, pp.\  1918--1923, 2007.

\bibitem[Huang et~al.(2017{\natexlab{a}})Huang, Papernot, Goodfellow, Duan, and Abbeel]{HuPaGoDuAb17}
S.~Huang, N.~Papernot, I.~Goodfellow, Y.~Duan, and P.~Abbeel.
\newblock {Adversarial Attacks on Neural Network Policies}, 2017{\natexlab{a}}.
\newblock Technical Report. \url{https://arxiv.org/abs/1702.02284}.

\bibitem[Huang et~al.(2017{\natexlab{b}})Huang, Kwiatkowska, Wang, and Wu]{HuKwWaWu17}
X.~Huang, M.~Kwiatkowska, S.~Wang, and M.~Wu.
\newblock {Safety Verification of Deep Neural Networks}.
\newblock In \emph{Proc. 29th Int. Conf. on Computer Aided Verification (CAV)}, pp.\  3--29, 2017{\natexlab{b}}.

\bibitem[Kamran et~al.(2022)Kamran, Sim{\~a}o, Yang, Ponnambalam, Fischer, Spaan, and Lauer]{kamran2022modern}
D.~Kamran, T.~Sim{\~a}o, Q.~Yang, C.~T Ponnambalam, J.~Fischer, M.~Spaan, and M.~Lauer.
\newblock {A Modern Perspective on Safe Automated Driving for Different Traffic Dynamics Using Constrained Reinforcement Learning}.
\newblock In \emph{2022 IEEE 25th Int. Conf. on Intelligent Transportation Systems (ITSC)}, pp.\  4017--4023. IEEE, 2022.

\bibitem[Karamzade et~al.(2024)Karamzade, Kim, Kalsi, and Fox]{karamzade2024reinforcement}
A.~Karamzade, K.~Kim, M.~Kalsi, and R.~Fox.
\newblock {Reinforcement Learning from Delayed Observations via World Models}.
\newblock \emph{arXiv preprint arXiv:2403.12309}, 2024.

\bibitem[Katz et~al.(2017)Katz, Barrett, Dill, Julian, and Kochenderfer]{KaBaDiJuKo17}
G.~Katz, C.~Barrett, D.~Dill, K.~Julian, and M.~Kochenderfer.
\newblock {Reluplex: An Efficient SMT Solver for Verifying Deep Neural Networks}.
\newblock In \emph{Proc. 29th Int. Conf. on Computer Aided Verification (CAV)}, pp.\  97--117, 2017.

\bibitem[Katz et~al.(2019{\natexlab{a}})Katz, Huang, Ibeling, Julian, Lazarus, Lim, Shah, Thakoor, Wu, Zelji\'c, Dill, Kochenderfer, and Barrett]{KaHuIbHuLaLiShThWuZeDiKoBa19}
G.~Katz, D.~Huang, D.~Ibeling, K.~Julian, C.~Lazarus, R.~Lim, P.~Shah, S.~Thakoor, H.~Wu, A.~Zelji\'c, D.~Dill, M.~Kochenderfer, and C.~Barrett.
\newblock {The Marabou Framework for Verification and Analysis of Deep Neural Networks}.
\newblock In \emph{Proc. 31st Int. Conf. on Computer Aided Verification (CAV)}, pp.\  443--452, 2019{\natexlab{a}}.

\bibitem[Katz et~al.(2019{\natexlab{b}})Katz, Huang, Ibeling, Julian, Lazarus, Lim, Shah, Thakoor, Wu, Zelji\'c, Dill, Kochenderfer, and Barrett]{KaHuIbJuLaLiShThWuZeDiKoBa19}
G.~Katz, D.~Huang, D.~Ibeling, K.~Julian, C.~Lazarus, R.~Lim, P.~Shah, S.~Thakoor, H.~Wu, A.~Zelji\'c, D.~Dill, M.~Kochenderfer, and C.~Barrett.
\newblock {The Marabou Framework for Verification and Analysis of Deep Neural Networks}.
\newblock In \emph{Proc. 31st Int. Conf. on Computer Aided Verification (CAV)}, pp.\  443--452, 2019{\natexlab{b}}.

\bibitem[Katz et~al.(2021)Katz, Barrett, Dill, Julian, and Kochenderfer]{KaBaDiJuKo21}
G.~Katz, C.~Barrett, D.~Dill, K.~Julian, and M.~Kochenderfer.
\newblock {Reluplex: a Calculus for Reasoning about Deep Neural Networks}.
\newblock \emph{Formal Methods in System Design (FMSD)}, 2021.

\bibitem[Kazak et~al.(2019)Kazak, Barrett, Katz, and Schapira]{KaBaKaSc19}
Y.~Kazak, C.~Barrett, G.~Katz, and M.~Schapira.
\newblock {Verifying Deep-RL-Driven Systems}.
\newblock In \emph{Proc. 1st ACM SIGCOMM Workshop on Network Meets AI \& ML (NetAI)}, pp.\  83--89, 2019.

\bibitem[Kirkpatrick et~al.(1983)Kirkpatrick, Gelatt, and Vecchi]{KiGeVe83}
S.~Kirkpatrick, C.~Gelatt, and M.~Vecchi.
\newblock {Optimization by Simulated Annealing}.
\newblock \emph{Science}, 220\penalty0 (4598):\penalty0 671--680, 1983.

\bibitem[Kober et~al.(2013)Kober, Bagnell, and Peters]{kober2013reinforcement}
J.~Kober, J.~Bagnell, and J.~Peters.
\newblock {Reinforcement Llearning in Robotics: A Survey}, 2013.

\bibitem[Kuper et~al.(2018)Kuper, Katz, Gottschlich, Julian, Barrett, and Kochenderfer]{KuKaGoJuBaKo18}
L.~Kuper, G.~Katz, J.~Gottschlich, K.~Julian, C.~Barrett, and M.~Kochenderfer.
\newblock {Toward Scalable Verification for Safety-Critical Deep Networks}, 2018.
\newblock Technical Report. \url{https://arxiv.org/abs/1801.05950}.

\bibitem[Ligatti et~al.(2009)Ligatti, Bauer, and Walker]{LiBauWal09}
J.~Ligatti, L.~Bauer, and D.~Walker.
\newblock {Run-Time Enforcement of Nonsafety Policies}.
\newblock \emph{{ACM} Trans. Inf. Syst. Secur.}, 12\penalty0 (3), 2009.

\bibitem[Liu et~al.(2019)Liu, Arnon, Lazarus, Barrett, and Kochenderfer]{LiArLaBaKo19}
C.~Liu, T.~Arnon, C.~Lazarus, C.~Barrett, and M.~Kochenderfer.
\newblock {Algorithms for Verifying Deep Neural Networks}, 2019.
\newblock Technical Report. \url{http://arxiv.org/abs/1903.06758}.

\bibitem[Liu et~al.(2020)Liu, Ding, and Liu]{LiDiLi20}
Y.~Liu, J.~Ding, and X.~Liu.
\newblock {IPO: Interior-Point Policy Optimization under Constraints}.
\newblock In \emph{Proc. 34th AAAI Conf. on Artificial Intelligence (AAAI)}, pp.\  4940--4947, 2020.

\bibitem[Lomuscio \& Maganti(2017)Lomuscio and Maganti]{LoMa17}
A.~Lomuscio and L.~Maganti.
\newblock {An Approach to Reachability Analysis for Feed-Forward ReLU Neural Networks}, 2017.
\newblock Technical Report. \url{http://arxiv.org/abs/1706.07351}.

\bibitem[Lyu et~al.(2020)Lyu, Ko, Kong, Wong, Lin, and Daniel]{LyKoKoWoLiDa20}
Z.~Lyu, C.~Y. Ko, Z.~Kong, N.~Wong, D.~Lin, and L.~Daniel.
\newblock {Fastened Crown: Tightened Neural Network Robustness Certificates}.
\newblock In \emph{Proc. 34th AAAI Conf. on Artificial Intelligence (AAAI)}, pp.\  5037--5044, 2020.

\bibitem[Ma et~al.(2024)Ma, Liu, Li, Zheng, Sun, and Chen]{ma2024learn}
H.~Ma, C.~Liu, S.~Li, S.~Zheng, W.~Sun, and J.~Chen.
\newblock {Learn Zero-Constraint-Violation Safe Policy in Model-Free Constrained Reinforcement Learning}.
\newblock \emph{IEEE Transactions on Neural Networks and Learning Systems}, 2024.

\bibitem[Ma et~al.(2020)Ma, Ding, and Mei]{MaDiMe20}
J.~Ma, S.~Ding, and Q.~Mei.
\newblock {Towards More Practical Adversarial Attacks on Graph Neural Networks}.
\newblock In \emph{Proc. 34th Conf. on Neural Information Processing Systems (NeurIPS)}, 2020.

\bibitem[Mandal et~al.(2024)Mandal, Amir, Wu, Daukantas, Newell, Ravaioli, Meng, Durling, Ganai, Shim, Katz, and Barrett]{MaAmWuDaNeRaMeDuGaShKaBa24}
U.~Mandal, G.~Amir, H.~Wu, I.~Daukantas, F.~Newell, U.~Ravaioli, B.~Meng, M.~Durling, M.~Ganai, T.~Shim, G.~Katz, and C.~Barrett.
\newblock {Formally Verifying Deep Reinforcement Learning Controllers with Lyapunov Barrier Certificates}, 2024.
\newblock Technical Report. \url{https://arxiv.org/abs/2405.14058}.

\bibitem[Manna \& Pnueli(1995)Manna and Pnueli]{MaPn95}
Z.~Manna and A.~Pnueli.
\newblock \emph{{Temporal Verification of Reactive Systems: Safety}}.
\newblock Springer Science \& Business Media, 1995.

\bibitem[Marchesini \& Farinelli(2020)Marchesini and Farinelli]{marchesini2020discrete}
E.~Marchesini and A.~Farinelli.
\newblock {Discrete Deep Reinforcement Learning for Mapless Navigation}.
\newblock In \emph{2020 IEEE International Conference on Robotics and Automation (ICRA)}, 2020.

\bibitem[Marchesini et~al.(2022)Marchesini, Corsi, and Farinelli]{marchesini2022exploring}
E.~Marchesini, D.~Corsi, and A.~Farinelli.
\newblock {Exploring Safer Behaviors for Deep Reinforcement Learning}.
\newblock In \emph{Proc. of the AAAI Conference on Artificial Intelligence}, 2022.

\bibitem[Marzari et~al.(2023)Marzari, Corsi, Marchesini, Farinelli, and Cicalese]{marzari2023enumerating}
L.~Marzari, D.~Corsi, E.~Marchesini, A.~Farinelli, and F.~Cicalese.
\newblock {Enumerating Safe Regions in Deep Neural Networks with Provable Probabilistic Guarantees}.
\newblock \emph{arXiv preprint arXiv:2308.09842}, 2023.

\bibitem[Mnih et~al.(2013)Mnih, Kavukcuoglu, Silver, Graves, Antonoglou, Wierstra, and Riedmiller]{MnKaSi13}
V.~Mnih, K.~Kavukcuoglu, D.~Silver, A.~Graves, I.~Antonoglou, D.~Wierstra, and M.~Riedmiller.
\newblock {Playing Atari with Deep Reinforcement Learning}, 2013.
\newblock Technical Report. \url{https://arxiv.org/abs/1312.5602}.

\bibitem[Monniaux(2008)]{Mon08}
David Monniaux.
\newblock {A Quantifier Elimination Algorithm for Linear Real Arithmetic}.
\newblock In \emph{Proc. of the 15th International Conference in Logic for Programming, Artificial Intelligence, and Reasoning ({LPAR} 2008)}, volume 5330 of \emph{LNCS}, pp.\  243--257. Springer, 2008.
\newblock \doi{10.1007/978-3-540-89439-1\_18}.
\newblock URL \url{https://doi.org/10.1007/978-3-540-89439-1\_18}.

\bibitem[Piterman et~al.(2006)Piterman, Pnueli, and Sa’ar]{PiPnSa06}
N.~Piterman, A.~Pnueli, and Y.~Sa’ar.
\newblock {Synthesis of Reactive (1) Designs}.
\newblock In \emph{Proc. 7th Int. Conf. on Verification, Model Checking, and Abstract Interpretation (VMCAI)}, pp.\  364--380, 2006.

\bibitem[Pnueli(1977)]{Pn77}
A.~Pnueli.
\newblock {The Temporal Logic of Programs}.
\newblock In \emph{Proc. of 18th Annual Symposium on Foundations of Computer Science (SFCS)}, pp.\  46--57, 1977.

\bibitem[Pore et~al.(2021)Pore, Corsi, Marchesini, Dall’Alba, Casals, Farinelli, and Fiorini]{pore2021safe}
A.~Pore, D.~Corsi, E.~Marchesini, D.~Dall’Alba, A.~Casals, A.~Farinelli, and P.~Fiorini.
\newblock {Safe Reinforcement Learning Using Formal Verification for Tissue Retraction in Autonomous Robotic-Assisted Surgery}.
\newblock In \emph{2021 IEEE/RSJ International Conference on Intelligent Robots and Systems (IROS)}, 2021.

\bibitem[Pranger et~al.(2021{\natexlab{a}})Pranger, K{\"{o}}nighofer, Posch, and Bloem]{PraKoPoBlo21}
S.~Pranger, B.~K{\"{o}}nighofer, L.~Posch, and R.~Bloem.
\newblock {{TEMPEST} - Synthesis Tool for Reactive Systems and Shields in Probabilistic Environments}.
\newblock In \emph{Proc. 19th Int. Symposium in Automated Technology for Verification and Analysis, (ATVA)}, volume 12971, pp.\  222--228, 2021{\natexlab{a}}.

\bibitem[Pranger et~al.(2021{\natexlab{b}})Pranger, K{\"{o}}nighofer, Tappler, Deixelberger, Jansen, and Bloem]{PraKoTa21}
S.~Pranger, B.~K{\"{o}}nighofer, M.~Tappler, M.~Deixelberger, N.~Jansen, and R.~Bloem.
\newblock {Adaptive Shielding under Uncertainty}.
\newblock In \emph{American Control Conference, (ACC)}, pp.\  3467--3474, 2021{\natexlab{b}}.

\bibitem[Ray et~al.(2019)Ray, Achiam, and Amodei]{RaAcAm19}
A.~Ray, J.~Achiam, and D.~Amodei.
\newblock {Benchmarking Safe Exploration in Deep Reinforcement Learning}, 2019.

\bibitem[Refaeli \& Katz(2022)Refaeli and Katz]{ReKa22}
I.~Refaeli and G.~Katz.
\newblock {Minimal Multi-Layer Modifications of Deep Neural Networks}.
\newblock In \emph{Proc. 5th Workshop on Formal Methods for ML-Enabled Autonomous Systems (FoMLAS)}, 2022.

\bibitem[Rodriguez \& S\'{a}nchez(2023)Rodriguez and S\'{a}nchez]{RoCe23a}
A.~Rodriguez and C.~S\'{a}nchez.
\newblock {Boolean Abstractions for Realizabilty Modulo Theories}.
\newblock In \emph{Proc. of the 35th Int. Conf. on Computer Aided Verification (CAV'23)}, 2023.

\bibitem[Rodriguez \& S\'{a}nchez(2024{\natexlab{a}})Rodriguez and S\'{a}nchez]{RoCe24}
A.~Rodriguez and C.~S\'{a}nchez.
\newblock {Adaptive Reactive Synthesis for LTL and LTLf Modulo Theories}.
\newblock In \emph{Proc. of the 38th AAAI Conf. on Artificial Intelligence}, pp.\  5037--5044, 2024{\natexlab{a}}.

\bibitem[Rodriguez \& S\'{a}nchez(2024{\natexlab{b}})Rodriguez and S\'{a}nchez]{RoCe24b}
A.~Rodriguez and C.~S\'{a}nchez.
\newblock Realizability modulo theories.
\newblock \emph{Journal of Logical and Algebraic Methods in Programming}, pp.\  100971, 2024{\natexlab{b}}.
\newblock ISSN 2352-2208.
\newblock \doi{https://doi.org/10.1016/j.jlamp.2024.100971}.

\bibitem[Rodriguez et~al.(2024)Rodriguez, Amir, Corsi, Sanchez, and Katz]{RoAmCoSaKa24}
A.~Rodriguez, G.~Amir, D.~Corsi, C.~Sanchez, and G.~Katz.
\newblock {Shield Synthesis for LTL Modulo Theories }, 2024.
\newblock Technical Report. \url{https://arxiv.org/abs/2406.04184}.

\bibitem[Rolf et~al.(2023)Rolf, Jackson, M{\"u}ller, Lang, Reggelin, and Ivanov]{rolf2023review}
B.~Rolf, I.~Jackson, M.~M{\"u}ller, S.~Lang, T.~Reggelin, and D.~Ivanov.
\newblock {A Review on Reinforcement Learning Algorithms and Applications in Supply Chain Management}, 2023.

\bibitem[Roy et~al.(2021)Roy, Girgis, Romoff, Bacon, and Pal]{RoGiRo21}
J.~Roy, R.~Girgis, J.~Romoff, P.~Bacon, and C.~Pal.
\newblock {Direct Behavior Specification via Constrained Reinforcement Learning}, 2021.
\newblock Technical Report. \url{https://arxiv.org/abs/2112.12228}.

\bibitem[Schneider(2000)]{Schne00}
F.~Schneider.
\newblock {Enforceable Security Policies}.
\newblock \emph{{ACM} Trans. Inf. Syst. Secur.}, 3\penalty0 (1):\penalty0 30--50, 2000.

\bibitem[Schulman et~al.(2017)Schulman, Wolski, Dhariwal, Radford, and Klimov]{ShWoDh17}
J.~Schulman, F.~Wolski, P.~Dhariwal, A.~Radford, and O.~Klimov.
\newblock {Proximal Policy Optimization Algorithms}, 2017.
\newblock Technical Report. \url{http://arxiv.org/abs/1707.06347}.

\bibitem[Sim{\~a}o et~al.(2021)Sim{\~a}o, Jansen, and Spaan]{simao2021alwayssafe}
T.~Sim{\~a}o, N.~Jansen, and M.~Spaan.
\newblock {AlwaysSafe: Reinforcement Learning without Safety Constraint Violations during Training}.
\newblock In \emph{Proc. of the 20th Int. Conf.e on Autonomous Agents and MultiAgent Systems (AAMAS)}, 2021.

\bibitem[Singh et~al.(2019)Singh, Gehr, Puschel, and Vechev]{SiGePuVe19}
G.~Singh, T.~Gehr, M.~Puschel, and M.~Vechev.
\newblock {An Abstract Domain for Certifying Neural Networks}.
\newblock In \emph{Proc. 46th ACM SIGPLAN Symposium on Principles of Programming Languages (POPL)}, 2019.

\bibitem[Singh et~al.(2022)Singh, Chen, Singhania, Nanavati, Gupta, et~al.]{singh2022reinforcement}
V.~Singh, S.~Chen, M.~Singhania, B.~Nanavati, A.~Gupta, et~al.
\newblock How are reinforcement learning and deep learning algorithms used for big data based decision making in financial industries--a review and research agenda.
\newblock \emph{International Journal of Information Management Data Insights}, 2022.

\bibitem[Srinivasan et~al.(2020)Srinivasan, Eysenbach, Ha, Tan, and Finn]{SrEyHa20}
K.~Srinivasan, B.~Eysenbach, S.~Ha, J.~Tan, and C.~Finn.
\newblock {Learning to be Safe: Deep RL with a Safety Critic}, 2020.
\newblock Technical Report. \url{https://arxiv.org/abs/2010.14603}.

\bibitem[Stooke et~al.(2020)Stooke, Achiam, and Abbeel]{StAcAb20}
A.~Stooke, J.~Achiam, and P.~Abbeel.
\newblock {Responsive Safety in Reinforcement Learning by Pid Lagrangian Methods}.
\newblock In \emph{Proc. 37th Int. Conf. on Machine Learning (ICML)}, pp.\  9133--9143, 2020.

\bibitem[Sun et~al.(2019)Sun, Khedr, and Shoukry]{SuKhSh19}
X.~Sun, H.~Khedr, and Y.~Shoukry.
\newblock {Formal Verification of Neural Network Controlled Autonomous Systems}.
\newblock In \emph{Proc. 22nd ACM Int. Conf. on Hybrid Systems: Computation and Control (HSCC)}, 2019.

\bibitem[Sutton \& Barto(2018)Sutton and Barto]{sutton2018reinforcement}
R.~Sutton and A.~Barto.
\newblock \emph{{Reinforcement Learning: An Introduction}}.
\newblock MIT Press, 2018.

\bibitem[Szegedy et~al.(2013)Szegedy, Zaremba, Sutskever, Bruna, Erhan, Goodfellow, and Fergus]{SzZaSuBrErGoFe13}
C.~Szegedy, W.~Zaremba, I.~Sutskever, J.~Bruna, D.~Erhan, I.~Goodfellow, and R.~Fergus.
\newblock {Intriguing Properties of Neural Networks}, 2013.
\newblock Technical Report. \url{http://arxiv.org/abs/1312.6199}.

\bibitem[Tai et~al.(2017)Tai, Paolo, and Liu]{TaPaLi17}
L.~Tai, G.~Paolo, and M.~Liu.
\newblock {Virtual-to-Real Deep Reinforcement Learning: Continuous Control of Mobile Robots for Mapless Navigation}.
\newblock In \emph{Proc. IEEE/RSJ Int. Conf. on Intelligent Robots and Systems (IROS)}, 2017.

\bibitem[Tessler et~al.(2019)Tessler, Mankowitz, and Mannor]{ChDaSh19}
C.~Tessler, D.~Mankowitz, and S.~Mannor.
\newblock {Reward Constrained Policy Optimization}.
\newblock In \emph{Proc. 7th Int. Conf. on Learning Representations (ICLR)}, 2019.

\bibitem[Thomas(2008)]{Th08}
W.~Thomas.
\newblock {Church’s Problem and a Tour Through Automata Theory}.
\newblock In \emph{Pillars of Computer Science}, pp.\  635--655. Springer, 2008.

\bibitem[Tjeng et~al.(2017)Tjeng, Xiao, and Tedrake]{TjXiTe17}
V.~Tjeng, K.~Xiao, and R.~Tedrake.
\newblock {Evaluating Robustness of Neural Networks with Mixed Integer Programming}, 2017.
\newblock Technical Report. \url{http://arxiv.org/abs/1711.07356}.

\bibitem[Vasi{\'c} et~al.(2022)Vasi{\'c}, Petrovi{\'c}, Wang, Nikoli{\'c}, Singh, and Khurshid]{VaPeWaNiSiKh22}
M.~Vasi{\'c}, A.~Petrovi{\'c}, K.~Wang, M.~Nikoli{\'c}, R.~Singh, and S.~Khurshid.
\newblock {Mo{\"E}T: Mixture of Expert Trees and its Application to Verifiable Reinforcement Learning}.
\newblock \emph{Neural Networks}, 151:\penalty0 34--47, 2022.

\bibitem[Wang et~al.(2018)Wang, Pei, Whitehouse, Yang, and Jana]{WaPeWhYaJa18}
S.~Wang, K.~Pei, J.~Whitehouse, J.~Yang, and S.~Jana.
\newblock {Formal Security Analysis of Neural Networks using Symbolic Intervals}.
\newblock In \emph{Proc. 27th {USENIX} Security Symposium}, pp.\  1599--1614, 2018.

\bibitem[Wolfgang(1990)]{Tho90}
T.~Wolfgang.
\newblock {Automata on Infinite Objects}.
\newblock In \emph{Handbook of Theoretical Computer Science, Volume {B:} Formal Models and Semantics}, pp.\  133--191. Elsevier and {MIT} Press, 1990.

\bibitem[Wu et~al.(2024)Wu, Isac, Zelji\'c, Tagomori, Daggitt, Kokke, Refaeli, Amir, Julian, Bassan, Huang, Lahav, Wu, Zhang, Komendantskaya, Katz, and Barrett]{WuIsZeTaDaKoReAmJuBaHuLaWuZhKoKaBa24}
H.~Wu, O.~Isac, A.~Zelji\'c, T.~Tagomori, M.~Daggitt, W.~Kokke, I.~Refaeli, G.~Amir, K.~Julian, S.~Bassan, P.~Huang, O.~Lahav, M.~Wu, M.~Zhang, E.~Komendantskaya, G.~Katz, and C.~Barrett.
\newblock {Marabou 2.0: A Versatile Formal Analyzer of Neural Networks}, 2024.

\bibitem[Wu et~al.(2019)Wu, Wang, Deshmukh, and Wang]{WuMaDeWa19}
M.~Wu, J.~Wang, J.~Deshmukh, and C.~Wang.
\newblock {Shield Synthesis for Real: Enforcing Safety in Cyber-Physical Systems}.
\newblock In \emph{Proc. 19th Int. Conf. on Formal Methods in Computer-Aided Design (FMCAD)}, pp.\  129--137, 2019.

\bibitem[Yang et~al.(2022{\natexlab{a}})Yang, Ji, Dai, Zhang, Zhou, Li, Yang, and Pan]{yang2022constrained}
L.~Yang, J.~Ji, J.~Dai, L.~Zhang, B.~Zhou, P.~Li, Y.~Yang, and G.~Pan.
\newblock {Constrained Update Projection Approach to Safe Policy Optimization}.
\newblock \emph{Advances in Neural Information Processing Systems (NeurIPS)}, 2022{\natexlab{a}}.

\bibitem[Yang et~al.(2022{\natexlab{b}})Yang, Sim{\~a}o, Jansen, Tindemans, and Spaan]{YaSiJa22}
Q.~Yang, T.~Sim{\~a}o, N.~Jansen, S.~Tindemans, and M.~Spaan.
\newblock {Training and Transferring Safe Policies in Reinforcement Learning}.
\newblock In \emph{AAMAS 2022 Workshop on Adaptive Learning Agents}, 2022{\natexlab{b}}.

\bibitem[Yerushalmi et~al.(2022)Yerushalmi, Amir, Elyasaf, Harel, Katz, and Marron]{YeAmElHaKaMa22}
R.~Yerushalmi, G.~Amir, A.~Elyasaf, D.~Harel, G.~Katz, and A.~Marron.
\newblock {Scenario-Assisted Deep Reinforcement Learning}.
\newblock In \emph{Proc. 10th Int. Conf. on Model-Driven Engineering and Software Development (MODELSWARD)}, pp.\  310--319, 2022.

\bibitem[Yerushalmi et~al.(2023)Yerushalmi, Amir, Elyasaf, Harel, Katz, and Marron]{YeAmElHaKaMa23}
R.~Yerushalmi, G.~Amir, A.~Elyasaf, D.~Harel, G.~Katz, and A.~Marron.
\newblock {Enhancing Deep Reinforcement Learning with Scenario-Based Modeling}.
\newblock \emph{SN Computer Science}, 4\penalty0 (2):\penalty0 156, 2023.

\bibitem[Zhang et~al.(2018)Zhang, Weng, Chen, Hsieh, and Daniel]{ZhWeChHsDa18}
H.~Zhang, T.~Weng, P.~Chen, C.~Hsieh, and L.~Daniel.
\newblock {Efficient Neural Network Robustness Certification with General Activation Functions}, 2018.

\bibitem[Zhang et~al.(2020)Zhang, Vuong, and Ross]{zhang2020first}
Y.~Zhang, Q.~Vuong, and K.~Ross.
\newblock {First Order Constrained Optimization in Policy Space}.
\newblock \emph{Proc. 34th Conf. on Neural Information Processing Systems (NeurIPS)}, 2020.

\end{thebibliography}
\bibliographystyle{rlc}

\clearpage \appendix

\section{Training Details}
\label{sec:app:training}

This appendix provides additional information regarding the described benchmarks, training setups, and safety requirements. Both analyzed benchmarks are navigation tasks with varying degrees of complexity and realism. 


\subsection{\gridenv{}}
Our first environment is depicted in Fig.~\ref{fig:motivation:training} (left). The goal of the agent (black square) is to reach the target position (yellow square) while avoiding collisions with walls and obstacles; the map is randomly selected from five different configurations. The agent does not have access to the full map of the environment but only to local information, and this makes this task an abstraction of the well-studied robotic \textit{mapless navigation} task~\citep{TaPaLi17, corsi2024analyzing}. The action and observation spaces are continuous and hence, the agent could move to any possible position in the arena; the agent is equipped with four proximity sensors that detect the distance from the closest obstacle in the respective directions, i.e., \texttt{Left, Right, Up, Down}. Although the action space is continuous, at each step, the agent is allowed to move only in one of the aforementioned directions, e.g., \textit{at time-step $t_0$ the agent performs a translation on the left of $0.321$ units}.

\paragraph{State/Action Spaces and DNN Topology.}
The structure of the neural network is inspired by recent work in the literature demonstrating that this task can be learned by a simple \textit{multi-layer perceptron} (MLP) encompassing relatively few nodes and hidden layers~\citep{marchesini2020discrete}. Next, we present a more detailed description of the MLP's structure:

\begin{itemize} \itemsep0em
    \item The \textit{input layer} constitutes $8$ neurons: the first $4$ neurons represent the distance from the closest obstacle in each direction, the following $2$ neurons encode the current position of the agent ($x$ and $y$ coordinates), while the last $2$ neurons encode the target's position. All these values are normalized in the interval $[0, 1]$ and can take on any continuous values within this interval.
    
    \item Two fully-connected \textit{hidden layers} of $16$ neurons each, with \textit{ReLU} activation functions.
    
    \item An \textit{output layer} of $4$ neurons, each representing the translation action in one of the possible directions (i.e., \texttt{Left, Right, Up, Down}); the values are continuous as the agent can translate any distance in a given direction. Crucially, the agent is constrained to move in only one direction at each time-step, hence, we always select the action with the highest value among the four options.
\end{itemize}

\paragraph{Training.}
We trained our agents with the state-of-the-art \emph{Proximal Policy Optimization} (PPO) algorithm~\citep{ShWoDh17}, which is widely considered the state-of-the-art. For this first task we employed a discrete reward function, that provides a positive reward when reaching the target and a negative reward for each collision; formally: 
\begin{equation}
\begin{aligned}
    R_t &= \begin{cases}
        +5 & \text{target reached} \\
        -1 & \text{collision with obstacle}
    \end{cases}
\end{aligned}
\end{equation}

where both conditions represent a terminal state. Following is a list of hyperparameters employed during training:
\begin{itemize} \itemsep-0.3em
    \item \textit{training episodes}: 500
    \item \textit{number of hidden layers}: 2
    \item \textit{size of hidden layers}: 16
    \item \textit{parallel environments}: 1
    \item \textit{gamma ($\gamma$)}: 0.995
    \item \textit{learning rate}: 0.0013
    \item \textit{memory limit}: None
    \item \textit{update frequency}: 4096 steps
    \item \textit{trajectory reduction strategy}: sum
    \item \textit{epochs}: 50
    \item \textit{batch number}: 64
    \item \textit{critic network size}: 2x256
    \item \textit{PPO clip}: 0.2
    \item \textit{GAE lambda}: 0.99
    \item \textit{target kl-divergence}: 0.02
    \item \textit{max gradient normal}: 0.5
\end{itemize}

\paragraph{Safety Requirements.} As mentioned, \gridenv{} is an abstraction of a real-world navigation problem; therefore, the crucial safety requirement is collision avoidance. Given the state and action space of the benchmark, the safety requirements involve only the first $4$ inputs (pertaining to the presence of obstacles) and the selected action. For more details regarding the encoding of the verification queries, see Appendix~\ref{sec:app:verification-properties}. From a high-level perspective, the safety requirements can be formalized as follows: \textit{``for any possible combination of agent and target position, the agent must not move towards an obstacle with a step-size larger than the distance to the closest obstacle in that direction"}.

\subsection{\mapless{}}
Our second environment is depicted in Fig.~\ref{fig:motivation:training} (right). Mapless navigation is a popular and well-studied task in the DRL literature~\citep{TaPaLi17, RaAcAm19, marchesini2020discrete}. This task is considered quite difficult due to the agent solely relying on local observations. For our experiments, we follow the same configuration presented in previous work in the field~\citep{pore2021safe, AmCoYeMaHaFaKa23}. In particular, the agent is equipped with a lidar sensor for obstacle detection, and with GPS and compass inputs for localization. 
A significant difference between \mapless{} and \gridenv{} is the degree of freedom for the agent. Specifically, in \mapless{}, the agent can simultaneously perform a linear step and a rotation, which provides additional movement options.

\paragraph{State/Action Spaces and DNN Topology.}
The structure of the neural network is similar to the one we employed for \gridenv{}, with the additional features derived from the sensors and actuators~\citep{RaAcAm19}. Following is a more detailed description of the structure:
\begin{itemize} \itemsep0em
    \item The \textit{input layer} constitutes of $9$ neurons, the first $7$ neurons represent lidar sensor readings, that indicate the distance from an obstacle in a given direction (from left to right, with a step of $30^{\circ}$). The final two input neurons indicate the target's position relative to the agent (i.e., polar coordinates of the target), calculated in real-time using GPS and compass data.
    
    \item $2$ fully-connected \textit{hidden layers} of $32$ neurons each, with \textit{ReLU} activation functions.
    \item An \textit{output layer} of $2$ neurons, the first neuron indicates the linear velocity (i.e., the speed of the robot), and the second one provides the angular velocity (i.e., a single value indicating the rotation). These two actions can be executed simultaneously, providing the agent with richer movement options.
\end{itemize}

\paragraph{Training.}
The training of agents on this benchmark was also based on PPO~\citep{ShWoDh17}. However, unlike the previous case, here we employed a \emph{continuous} reward function, given the increased complexity of this task: 
\begin{equation}
    R_t = \begin{cases}
    1 & \text{the goal is reached} \\
    -1 & \text{the agent collides} \\
    (dist_{t-1} - dist_{t}) \cdot \eta - \beta & \text{otherwise}\\
    \end{cases}
\end{equation}

where $dist_k$ is the distance from the target at time-step $k$; $\eta$ is a normalization factor; and $\beta$ is a penalty, intended to encourage the robot to reach the target quickly (in our experiments, we empirically set $\eta=3$ and $\beta=0.001$). Following is a list of hyperparameters employed during training:
\begin{itemize} \itemsep-0.3em
    \item \textit{training episodes}: 500
    \item \textit{number of hidden layers}: 2
    \item \textit{size of hidden layers}: 32
    \item \textit{parallel environments}: 1
    \item \textit{gamma ($\gamma$)}: 0.99
    \item \textit{learning rate}: 0.0003
    \item \textit{memory limit}: None
    \item \textit{update frequency}: 1024 steps
    \item \textit{trajectory reduction strategy}: sum
    \item \textit{epochs}: 10
    \item \textit{batch number}: 32
    \item \textit{critic network size}: 2x256
    \item \textit{PPO clip}: 0.2
    \item \textit{GAE lambda}: 0.95
    \item \textit{target kl-divergence}: 0.02
    \item \textit{max gradient normal}: 0.5
\end{itemize}

\paragraph{Safety Requirements.} The safety requirements for \mapless{} aim at guaranteeing the same objectives as the ones described for the \gridenv{} environment. However, there is a crucial difference in this context: the consequences of an action may not always be predictable because the agent's increased degree of freedom results in a set of possible collision situations that cannot be detected by observation alone. For example, there may be an obstacle between two lidar scans that the agent cannot detect. Therefore, we cannot ensure the safety of the agent in \emph{any} possible configuration, even if it meets all requirements. Our objective is thus to guarantee the agent's safety against the specified set of requirements, which may not encompass all potential collisions.

\section{Reactive Synthesis}
\label{sec:app:synthesis}


In continuation to the temporal operators described in Sec.~\ref{sec:preliminaries},
additional temporal operators include $\calR$ (\emph{release}), $\Event$(\emph{finally}),
and $\mathcal{W}$ (\emph{weak until}) which can also be derived from the recursive syntax, e.g., $\varphi_0 \calR \varphi_1 \equiv \neg (\neg \varphi_0 \U \neg \varphi_0)$.
In addition, we note that equivalences are also well defined, i.e., distribuitivity properties (e.g., $\Always (\varphi_0 \wedge \varphi_1) \equiv (\Always \varphi_0) \wedge (\Always \varphi_1)$), negation properties (e.g., $ \neg \Event \varphi \equiv \Always \neg \varphi$) and other temporal-specific properties (e.g., $\Always \varphi \equiv \Always \Always \varphi$).

 Let $\omega$ denote infinite words~\citep{Tho90},
then the semantics of LTL formulas associates traces $\sigma\in\Sigma^\omega$ with
LTL fomulae (where $\sigma \models \top$ always holds, and $\Or$ and $\neg$ are standard):
\[
  \begin{array}{l@{\hspace{0.3em}}c@{\hspace{0.3em}}l}
    \sigma \models a & \text{iff } & a \in\sigma(0) \\
     \sigma \models \Next \varphi & \text{iff } & \sigma^1\models \varphi \\
     \sigma \models \varphi_1 \U \varphi_2 & \text{iff } & \text{for some } i\geq 0\;\; \sigma^i\models \varphi_2, \text{ and } \text{for all } 0\leq j<i, \sigma^j\models\varphi_1 \\
  \end{array}
\]
A safety formula $\varphi$ is such that for every failing trace $\sigma\not\models\varphi$ there is a finite prefix $u$ of $\sigma$, such that all $\sigma'$ extending $u$ also falsify $\varphi$, i.e., $\sigma'\not\models\varphi$.
In this paper, we only synthesize models for safety formulae, which are indeed the most interesting ones for our problem and the fully monitorable ones.

Reactive LTL synthesis~\citep{PiPnSa06,Th08}
is the task of producing a system that satisfies a given LTL specification $\varphi$, where
atomic propositions in $\varphi$ are split into variables
controlled by the environment (``input variables'') and by the system (``output variables'').
Synthesis corresponds to a game where, in each turn, the
environment player produces values of the input propositions, and the system player
responds with values of the output propositions.
A play is an infinite sequence of turns, i.e., an infinite interaction of the system with the environment.
%
%
A strategy  for the system is said to be \emph{winning}
for the system if all the possible plays played
according to the strategy satisfy the LTL formula $\varphi$.

\section{Symbolic Representation: Additional Details}
\label{sec:app:symbolic}

In \emph{fully observable} domains, it is possible to encode the environment with symbolic representations.
 This includes a representation of the arena as a formula $\psi$ in propositional logic, which implies that its states can be precisely characterized as models of $\psi$. In other words, models $M=\{...,m_k,...\} \neq \emptyset$ of $\psi$ is a precise encoding of the arena, which means that each model can be obtained by performing classic Boolean satisfiability (SAT) queries over $\psi$,
 i.e., there exists a SAT encoding of the set of states.

 As a toy example, let us consider an arena with four states that correspond to coordinates north-south and west-east: $\{\text{NE},\text{NW},\text{SE},\text{SW}\}$. A formula $\phi = \text{NE} \vee \text{NW} \vee \text{SE} \vee \text{SW}$ encodes the whole arena, and a model of the formula is a concrete state.
 Moreover, we can encode groups of states of the arena using conjunctions.
 In the example above, the \textit{north} group is encoded precisely by $\phi = \psi \wedge \neg (\text{SE} \vee \text{SW})$.
 These observations are very relevant for verification-guided shielding due to our ability to use such symbolic representations in order to precisely encode and represent disjunct safe and unsafe regions.
 . 
 
 Note that, since we are in a continuous domain, the amount of states is infinite, so this representation is not encoded with propositional logic, but rather with first-order logic modulo appropriate theories (in our case, linear real arithmetic~\citep{Mon08}). Thus, we can obtain models from satisfiability modulo theory (SMT) queries, i.e.,
there is an SMT encoding of the set of states.
 We can modify the example above to show the difference. Consider the state is characterized by two input values of the environment, $x_1$ and $x_2$: \textit{south} is represented by $x_1: [0,1)$ (respectively, $x_1: [1,2]$ represents \textit{north}) and \textit{west} is represented by $x_2: [0,1)$ (respectively, $x_2: [1,2]$ represents \textit{east}).
 Then, the formula $\exists x_1,x_2. (0 \leq x_1 \leq 2) \wedge (0 \leq x_2 \leq 2) $ encodes all the infinite states in a succinct manner, i.e., states are models of this formula. Again, we can encode groups easily, e.g., it is possible to represent \textit{south-east} with models of $\exists x_1,x_2. (0 \leq x_1 < 1) \wedge (1 \leq x_2 \leq 2)$, etc.

 In summary, with symbolic encodings, we can compactly represent states (or regions) and sets of states and also simplify them.
 Note that these encodings are especially succinct if states have overlapping regions, i.e., share models in the SMT formula, since this allows the formula to be further simplified. 
 In our empirical evaluation, we used Z3's~\citep{DeBj08} \texttt{simplify(phi)} primitive for this step.
\section{Verification Example and Property Encodings}
\label{sec:app:verification-properties}

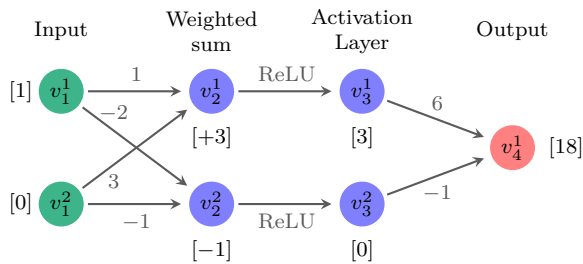
\begin{figure}[ht]
    \begin{center}
    \def\layersep{2.0cm}
    \begin{tikzpicture}[shorten >=1pt,->,draw=black!50, node
    distance=\layersep,font=\footnotesize]
    
    \node[input neuron] (I-1) at (0,-1) {$v^1_1$};
    \node[input neuron] (I-2) at (0,-2.5) {$v^2_1$};
    
    \node[left=-0.05cm of I-1] (b1) {$[1]$};
    \node[left=-0.05cm of I-2] (b2) {$[0]$};
    \node[hidden neuron] (H-1) at (\layersep,-1) {$v^1_2$};
    \node[hidden neuron] (H-2) at (\layersep,-2.5) {$v^2_2$};	
    \node[hidden neuron] (H-3) at (2*\layersep,-1) {$v^1_3$};
    \node[hidden neuron] (H-4) at (2*\layersep,-2.5) {$v^2_3$};
    \node[output neuron] at (3*\layersep, -1.75) (O-1) {$v^1_4$};
    
    \draw[nnedge] (I-1) --node[above] {$1$} (H-1);
    \draw[nnedge] (I-1) --node[above, pos=0.3] {$-2$} (H-2);
    \draw[nnedge] (I-2) --node[below, pos=0.3] {$3$} (H-1);
    \draw[nnedge] (I-2) --node[below] {$-1$} (H-2);
    \draw[nnedge] (H-1) --node[above] {ReLU} (H-3);
    \draw[nnedge] (H-2) --node[below] {ReLU} (H-4);
    \draw[nnedge] (H-3) --node[above] {$6$} (O-1);
    \draw[nnedge] (H-4) --node[below] {$-1$} (O-1);
    
    \node[below=0.05cm of H-1] (b1) {$[+3]$};
    \node[below=0.05cm of H-2] (b2) {$[-1]$};
    
    \node[below=0.05cm of H-3] (b1) {$[3]$};
    \node[below=0.05cm of H-4] (b2) {$[0]$};
    \node[right=0.05cm of O-1] (b1) {$[18]$};
    
    \node[annot,above of=H-1, node distance=0.8cm] (hl1) {Weighted sum};
    \node[annot,above of=H-3, node distance=0.8cm] (hl2) {Activation Layer };
    \node[annot,left of=hl1] {Input };
    \node[annot,right of=hl2] {Output };
    \end{tikzpicture}
    \caption{A toy DNN.}
    \label{fig:add:toynet}
    \end{center}
\end{figure}

Suppose we wish to verify that the toy DNN depicted in
Fig.~\ref{fig:add:toynet} outputs, for any given input, a value strictly larger than
$30$, i.e., for any input $x=\langle v_1^1,v_1^2\rangle$, the property
$N(x)=v_4^1 > 30$ always holds. 
It is straightforward to encode this property as a verification
query by using a precondition that does not restrict the inputs,
i.e., $P=(true)$, and also, by setting $Q=(v_4^1\leq 30)$ as a postcondition.
Hence, for the verification query $P(x_0) \wedge Q(N(x_0))$, a sound verification engine will return
\sat, along with a feasible counterexample, e.g.,
$x=\langle 1, 0\rangle$, which produces $v_4^1=24 \leq 30$. Hence, proving
that this property does not hold.
	
In this work, we used \emph{Marabou}~\citep{KaHuIbHuLaLiShThWuZeDiKoBa19, WuIsZeTaDaKoReAmJuBaHuLaWuZhKoKaBa24} as our verification engine. 
Marabou is sound and complete and has recently been used in various
applications~\citep{ElGoKa20, AmScKa21,AmWuBaKa21, AmZeKaSc22, ReKa22, ElCoKa22, AmFrKaMaRe23, AmMaZeKaSc23}.

\paragraph{Note.}
We note that similarly to previous work~\citep{AmCoYeMaHaFaKa23}, we typically considered violations of the required property, if the ``wrong'' action won by a given margin.

\subsection{Verification Queries for \gridenv{}}
The general idea behind these properties is to ensure that if an obstacle is detected in one of the $4$ possible directions, the agent will not take a step in that direction, and have the step size greater than the measured distance. The operator \texttt{argmax} encodes the fact that the agent can only move in one direction at a time, i.e., the one with the highest value; the constant $0.055$ indicates the linear speed of the agent, which should be multiplied by the DNN's output action to obtain the actual step size. For example, if the DNN outputs $Y=[0.3, 0.2, 0.8, 0.0]$, the agent will move \texttt{Up} (i.e., the action associated to the third node) by $0.8 \cdot 0.055 = 0.044$ units. Below we report the complete formalization of the properties, where $X$ is the input, $Y$ is the output, $\mathcal{D}_x$ is the domain of the input space, and $\mathcal{N}$ is the neural network function; all inputs are normalized to the interval $[0, 1]$. Finally, we note that if the expression returns \texttt{true} (\sat{}), it means that there is an assignment that violates the properties, and the network is deemed unsafe.

\begin{itemize}

    \item \texttt{\gridenv{} 1 (G1)}: avoid collision with an obstacle on the \textbf{right} of the agent.
    \begin{itemize}
        \item (\texttt{argmax}(Y)$==0$) and ($Y[0] \cdot 0.055 > X[0])$ and ($Y=\mathcal{N}(X)$) \hspace{15pt} $\forall_{X} \in \mathcal{D}_x$
    \end{itemize} 

    \item \texttt{\gridenv{} 2 (G2)}: avoid collision with an obstacle on the \textbf{left} of the agent.
    \begin{itemize}
        \item (\texttt{argmax}(Y)$==1$) and ($Y[1] \cdot 0.055 > X[1])$ and ($Y=\mathcal{N}(X)$) \hspace{15pt} $\forall_{X} \in \mathcal{D}_x$ 
    \end{itemize} 

    \item \texttt{\gridenv{} 3 (G3)}: avoid collision with an obstacle \textbf{above} the agent.
    \begin{itemize}
        \item (\texttt{argmax}(Y)$==2$) and ($Y[2] \cdot 0.055 > X[2])$ and ($Y=\mathcal{N}(X)$) \hspace{15pt} $\forall_{X} \in \mathcal{D}_x$ 
    \end{itemize} 

    \item \texttt{\gridenv{} 4 (G5)}: avoid collision with an obstacle \textbf{below} the agent.
    \begin{itemize}
        \item (\texttt{argmax}(Y)$==3$) and ($Y[3] \cdot 0.055 > X[3])$ and ($Y=\mathcal{N}(X)$) \hspace{15pt} $\forall_{X} \in \mathcal{D}_x$ 
    \end{itemize} 

\end{itemize}

\subsection{Verification Queries for \mapless{}}
The properties for the \mapless{} environment follow the same structure as explained for the previous benchmark. A crucial difference to note, which we already discussed in Appendix~\ref{sec:app:training}, is that given the complex nature of the problem and the agent's high degree of freedom, we cannot guarantee the \emph{absolute} safety of the agent. Hence, in this scenario, we settle instead on guaranteeing adherence to the following set of constraints.

\begin{itemize}

    \item \texttt{\mapless{} 1 (M1)}: avoid collision with an obstacle \textbf{in front} of the robot.
    \begin{itemize}
        \item ($X[3]-0.17 < Y[0] \cdot 0.015$) and ($Y=\mathcal{N}(X)$) \hspace{15pt} $\forall_{X} \in \mathcal{D}_x$  
    \end{itemize}

    \item \texttt{\mapless{} 2 (M2)}: avoid collision with an obstacle on the \textbf{left} of the robot.
    \begin{itemize}
        \item ($X[1]-0.17 < Y[0] \cdot 0.015$) and ($Y[1] < -0.2$) and ($Y=\mathcal{N}(X)$) \hspace{15pt} $\forall_{X} \in \mathcal{D}_x$
    \end{itemize}

    \item \texttt{\mapless{} 3 (M3)}: avoid collision with an obstacle \textit{slightly} on the \textbf{left} of the robot.
    \begin{itemize}
        \item ($X[2]-0.17 < Y[0] \cdot 0.015$) and ($Y[1] < -0.15$) and ($Y=\mathcal{N}(X)$) \hspace{15pt} $\forall_{X} \in \mathcal{D}_x$ 
    \end{itemize}

    \item \texttt{\mapless{} 4 (M4)}: avoid collision with an obstacle on the \textbf{right} of the robot.
    \begin{itemize}
        \item ($X[5]-0.17 < Y[0] \cdot 0.015$) and ($Y[1] > 0.2$) and ($Y=\mathcal{N}(X)$) \hspace{15pt} $\forall_{X} \in \mathcal{D}_x$ 
    \end{itemize}

    \item \texttt{\mapless{} 5 (M5)}: avoid collision with an obstacle \textit{slightly} on the \textbf{right} of the robot.
    \begin{itemize}
        \item ($X[4]-0.17 < Y[0] \cdot 0.015$) and ($Y[1] > 0.15$) and ($Y=\mathcal{N}(X)$) \hspace{15pt} $\forall_{X} \in \mathcal{D}_x$   
    \end{itemize} 
    
\end{itemize}
    

\section{Formal Verification of Safe Regions}
\label{sec:app:verification-code}
Algorithm \ref{alg:app:formal-validation} reports the pseudocode for step (2) of our approach, as described in Sec.~\ref{sec:method}. This subprocedure takes as input the approximated set of \sat{} regions $\tilde{S}$ (i.e., validated unsafe regions), and the approximated set of \unsat{} regions $\tilde{U}$ (i.e., potentially safe regions). The algorithm iterates over $\tilde{U}$, while verifying the regions with a formal verification tool (e.g., \texttt{Marabou}~\citep{KaHuIbHuLaLiShThWuZeDiKoBa19}), to formally ensure that these regions are actually safe. If the result is \sat{}, we relabel the region in question as unsafe. After the process, all regions in the \unsat{} set are formally safe. This ensures that decisions made by the policy in these regions are reliable without the need for shielding.

\begin{algorithm}[H]
\begin{algorithmic}[1]
\Require $\mathcal{\tilde{S}}$, $\mathcal{\tilde{U}}$
\Ensure $\mathcal{S}$, $\mathcal{U}$
\State $y \gets 1$, $\mathcal{S} \gets \emptyset$, $\mathcal{U} \gets \emptyset$ 
\For{region \textbf{in} $\mathcal{\tilde{U}}$}
\If{\texttt{formal-verificaiton(}region\texttt{)} \textbf{is} \sat{}} \Comment{call a verification tool backend}
    \State \texttt{remove(}region, $\mathcal{\tilde{U}}$\texttt{)}
    \State \texttt{add(}region, $\mathcal{S}$\texttt{)}
\EndIf
\State $\mathcal{U} \gets \mathcal{\tilde{U}}$, $\mathcal{S} \gets \mathcal{\tilde{S}} \cup \mathcal{S}$
\EndFor
\end{algorithmic}
\caption{Formal verification of safe regions.}
\label{alg:app:formal-validation}
\end{algorithm}

\end{document}